\documentclass{article}

\usepackage{microtype}
\usepackage{graphicx}
\usepackage{subfigure}
\usepackage{booktabs} 

\usepackage[hyphens]{url}
\usepackage{hyperref}
\hypersetup{colorlinks=true,breaklinks=true}

\usepackage[accepted]{icml2025}

\usepackage{amsmath}
\usepackage{amssymb}
\usepackage{mathtools}
\usepackage{amsthm}

\usepackage[capitalize,noabbrev]{cleveref}

\theoremstyle{plain}

\theoremstyle{definition}

\theoremstyle{remark}

\usepackage[textsize=tiny]{todonotes}

\usepackage{caption}
\DeclareCaptionFormat{cont}{#1 (Continued)#2#3\par}

\icmltitlerunning{Unveiling Reasoning Thresholds in Language Models: Scaling, Fine-Tuning, and Interpretability through Attention Maps}

\begin{document}

\twocolumn[
\icmltitle{Unveiling Reasoning Thresholds in Language Models: Scaling, Fine-Tuning, and Interpretability through Attention Maps}



\icmlsetsymbol{equal}{*}

\begin{icmlauthorlist}
\icmlauthor{Yen-Che Hsiao}{yyy}
\icmlauthor{Abhishek Dutta}{yyy}
\end{icmlauthorlist}

\icmlaffiliation{yyy}{Department of Electrical and Computer Engineering, University of Connecticut, Storrs CT 06269, USA}

\icmlcorrespondingauthor{Yen-Che Hsiao}{yen-che.hsiao@uconn.edu}

\icmlkeywords{Machine Learning, ICML}

\vskip 0.3in
]



\printAffiliationsAndNotice{}  

\begin{abstract}
This study investigates the in-context learning capabilities of various decoder-only transformer-based language models with different model sizes and training data, including GPT2, SmolLM2, OpenELM, TinyLlama, Stable LM, and Gemma 2. We identify a critical parameter threshold (~1.6 billion), beyond which reasoning performance improves significantly in tasks such as commonsense reasoning in multiple-choice
question answering and deductive reasoning. Specifically, models above this threshold achieve better success rates in chain-of-thought (CoT) prompting for deductive reasoning tasks, especially those requiring longer reasoning chains, such as proof by contradiction and disjunction elimination. To address limitations in sub-threshold models, we demonstrate that fine-tuning with task-specific exemplars substantially enhances reasoning performance, enabling accurate CoT generation even without additional exemplars in the prompt for tasks with shorter reasoning chains. Finally, our analysis of attention maps reveals that models capable of generating correct CoTs exhibit higher token-level attention scores on subsequent correct tokens and the correct parts of speech, providing interpretability insights into reasoning processes. These findings collectively advance understanding of reasoning capabilities in decoder-only transformer-based models. The code is available at: \url{https://github.com/AnnonymousForPapers/CoT_Reasoning_Test}.
\end{abstract}

\section{Introduction}

\citet{NEURIPS2020_1457c0d6} introduced GPT-3, a 175-billion-parameter language model (LM), and demonstrated that increasing the number of its parameter from 0.1 to 175 billion improves performance across 42 benchmarks, except for their reversed words task, the Word-in-Context task in the SuperGLUE benchmark \citep{NEURIPS2019_4496bf24}, and rounds 1 and 2 of the Adversarial Natural Language Inference dataset \citep{nie2020adversarialnlinewbenchmark}, by giving a few task demonstrations within the prompt at inference time without parameter updates, a technique commonly referred to as in-context learning (ICL) \citep{akyurek2022learning, min2021metaicl}.
\citet{wei2022chain} proposed chain-of-thought (CoT) prompting and showed that scaling up model size improved the performance of ICL and CoT prompting led to further gains on the models from PaLM \cite{chowdhery2022palmscalinglanguagemodeling}, LaMDA \cite{thoppilan2022lamda}, and GPT-3 \cite{NEURIPS2020_1457c0d6}, with improvements appearing to be the largest for PaLM 540B \cite{chowdhery2022palmscalinglanguagemodeling}.

\citet{shi2023language} evaluates the reasoning abilities of the 6B, 62B, and 540B models from PaLM \cite{chowdhery2022palmscalinglanguagemodeling} and different models from GPT-3 \cite{NEURIPS2020_1457c0d6} in multilingual settings and found that the accuracy increases as the number of parameters increases in the models from PaLM \cite{chowdhery2022palmscalinglanguagemodeling} on their proposed Multilingual Grade School Math benchmark across 11 different languages through CoT prompting \citep{wei2022chain}. \citet{zhou2024can} show that LMs (GPT-3.5-turbo-0613 \citep{floridi2020gpt}, Gemini-Pro (Jan. 2024) \citep{chowdhery2022palmscalinglanguagemodeling},  Llama2-70B \citep{touvron2023llama2openfoundation}, and Mixtral-8x7B \citep{jiang2024mixtralexperts}) can be distracted by irrelevant rationales that are unhelpful for solving a given question in the CoT exemplars. \citet{sia2024where} identify the layers in which "task recognition" occurs using causal
masking over different parts of the context and conduct exploratory studies into the extent to which subsequent layers are either redundant or corresponding to the "task recognition" layers on machine translation and code generation tasks. \citet{wibisono2024from} show that ICL for word analogy completion on a frequently co-occurring word pair can arise by modeling word co-occurrence using the continuous bag-of-words model \citep{mikolov2013efficient} trained on frequently co-occurring tokens in the training dataset, without needing positional information or attention mechanisms, they find out that positional information is essential when the ICL task is to predict the first token in a sentence, and they find out that their designed transformer with learned positional embedding results in higher accuracy for ICL to complete token pairs compared to sinusoidal or rotary positional encoding \citep{su2024roformer} if there are disturbing tokens between the word pairs in the exemplars. \citet{stechly2024chain} show that the success rate of solving procedural reasoning tasks using their considered LMs (GPT-4 \citep{openai2024gpt4technicalreport}, Claude-3-Opus \citep{anthropic2023claude3}, and GPT-4-Turbo) prompted by the CoT techniques \citep{wei2022chain} decreases as the generality of the prompt increases regardless of the number of sub-goals and the success rate also decreases as the number of sub-goals increases, regardless of the specificity of the CoT prompt.

We extend the work of \citet{wei2022chain} by investigating the reasoning abilities of 23 open-source, well-documented decoder-only transformer-based LMs through CoT prompting \citep{wei2022chain}. Focusing on models with fewer than 10 billion parameters, we identify a critical parameter gap separating models with reasoning ability from those without. Testing on the CommonsenseQA (CSQA) dataset \citep{talmor-etal-2019-commonsenseqa}, we observe a significant improvement in success rates for models with more than 1.6 billion parameters compared to those with fewer than 1.5 billion, except for LMs from the OpenELM family \citep{mehta2024openelm}. To minimize the influence of pre-training biases, we evaluate the models on the Proof and Ontology-Generated Question-Answering-Out-Of-Demonstration (PrOntoQA-OOD) dataset \citep{saparov2023language} using six distinct deductive rules. Notably, we find parameter gaps of 774 million to 1.1 billion parameters for disjunction elimination and 1.5 billion to 1.6 billion parameters for proof by contradiction.

To address the limitations of sub-threshold models, we fine-tune six of the smallest LMs on exemplars generated by the PrOntoQA-OOD dataset \citep{saparov2023language}. Fine-tuning enables these models to achieve significantly improved success rates on four deductive rules—implication elimination, conjunction introduction, conjunction elimination, and disjunction introduction—even without exemplars in the prompt. Finally, we analyze attention maps of the largest and smallest models. Our findings show that the largest model (Gemma2 9B IT) capable of generating correct CoTs exhibits higher token-level scores \citep{kang-shin-2023-samrank} for subsequent correct tokens and tokens corresponding to the correct parts of speech, in contrast to the smallest model (GPT2), which fails to generate accurate CoTs. In conclusion, this study establishes three key contributions: the identification of a critical parameter threshold necessary for reasoning abilities, the demonstrated efficacy of fine-tuning in enhancing the performance of models below this threshold, and the utilization of attention maps to provide interpretive insights into performance disparities across models of different sizes and reasoning capacities.

\section{Approaches}


Given a test question $x_{test}$ and an LM $f_\theta$, we expect to get the correct answer $y_{test}$ from the output of $f_\theta$. Following the definition in \citep{sia2024where}, when using the ICL techniques, the LM is prompted with the input $x_{ICL} = [S_n,x_{test}] = [x_1,y_1,\dots,x_n,y_n,x_{test}]$, which contains $n$ exemplars $S_n=\{(x_i,y_i)\}^n_{i=1}$, each consists of a question $x_i$ and answer $y_i$, and a given test question $x_{test}$. When using the CoT techniques \citep{wei2022chain}, the LM is prompted with the input $x_{CoT} = [x_1,\mathcal{T}_1,y_1,\dots,x_n,\mathcal{T}_n,y_n,x_{test}]$, where $\mathcal{T}_i=[\mathcal{T}^{(1)}_i, \mathcal{T}^{(2)}_i, \mathcal{T}^{(3)}_i, \dots, \mathcal{T}^{(k)}_i]$ is a step-by-step rationale which consists of several thoughts $\mathcal{T}^{(j)}_i$.

\subsection{CSQA dataset}

CommonsenseQA \cite{talmor-etal-2019-commonsenseqa} is a commonsense question answering dataset that contains 12,247 multiple-choice questions created by crowdsourcing workers using the knowledge encoded in CONCEPTNET \cite{speer2017conceptnet}. We use the validation set of the CommonsenseQA dataset \cite{talmor-etal-2019-commonsenseqa} which contains 1,221 multiple-choice questions. 

\subsubsection{Input prompt for the CoT experiment using CSQA dataset}

The prompt for the CoT experiment on the CSQA dataset \cite{talmor-etal-2019-commonsenseqa} contains a set of exemplars, a test question, and the choices for the test question. The exemplars are adopted from \citep{wei2022chain} as shown in Figure \ref{fig:CSQA_exemplars}. The first part of the input prompt for the CSQA experiments contains 7 sets of questions, gold CoT, and answers as shown in Figure \ref{fig:CSQA_exemplars}. In the initial prompt from \citep{wei2022chain}, each question is appended after "Q: ", the first answer choice is appended after " Answer Choices: (a) ", and the second to the last answer choice is appended after "\textbackslash n(" concatenated with the lower case of the answer key (from 'A' to 'E') and ") ", and the gold CoT is appended after "\textbackslash n A: ". In the later part of the input prompt, one of the questions is chosen from the validation set of the CSQA dataset \cite{talmor-etal-2019-commonsenseqa} without repetition and the text of the question is appended after "\textbackslash n Q: ", which is appended after the initial prompt in Figure \ref{fig:CSQA_exemplars}. The answer choices of the chosen question are appended after the chosen question in the same way as the prompt contains the 7 exemplars. "\textbackslash n" is appended after the text contains the last answer choice of the chosen question.


\subsubsection{String parsing strategy for the CoT experiments using CSQA dataset}
\label{String_parse_CSQA}

In the CSQA dataset, each $y_{test}$ is a string that contains "So the answer is (" concatenated with either "a", "b", "c", "d", or "e", correspond to the answer of the test question, $x_{test}$, and with ").". We aim to extract the answer from the output of the LM using string parsing.


For the CoT experiment, we first get the output of the LM, $f_\theta(x_{CoT})$, given the CoT prompt $x_{CoT}$. The output of the LM is a string which can be represented as $f_\theta(x_{CoT}) = [x_{CoT}, \mathcal{T}_{respond}, y_{response}, y_{else}]$, where $y_{response}$ is a string that contains the answer to the test question $x_{test}$, $\mathcal{T}_{response}$ is the thought for the answer, and $y_{else}$ is a string that contains content irrelevant to the test question. 

In our string parsing process, we first remove $x_{CoT}$ from $f_\theta(x_{CoT})$ and then extract the string after "So the answer is (" and before ")". If the extracted string is a character of either "a", "b", "c", "d", or "e", we compare the extracted string with the correct answer key, which is a character of either "A", "B", "C", "D", or "E", in the dataset. Regardless of capitalization, if the extracted string and the correct answer key are the same letter, the LM is considered to have answered the question, $x_{test}$, correctly; otherwise, the answer is considered incorrect. If the extracted string is not one of the five characters, the LM is considered to have not answered the question.

\subsection{PrOntoQA-OOD dataset}

PrOntoQA dataset \citep{saparov2023language} is a synthetic and programmable question-answering dataset that contains examples generated from a synthetic world model represented in first-order logic. PrOntoQA-OOD dataset \citep{saparov2023testing} is a synthetic and programmable reasoning dataset that extends the PrOntoQA dataset \citep{saparov2023language} to a complete set of deduction rules and to compositional proofs. Each example in PrOntoQA dataset \citep{saparov2023language} contains texts with a set of premises, a conclusion that is the target fact to be proved or disproved, and a gold CoT containing the proof of the conclusion. The six deduction rules include implication elimination (also known as modus ponens), conjunction introduction, conjunction elimination, disjunction introduction, disjunction elimination (also called proof by cases), and proof by contradiction.

\subsubsection{Input prompt for the CoT experiment using the PrOntoQA-OOD dataset}

The input prompt consists of several examples generated from the PrOntoQA-OOD dataset appended with the test question which is taken from the next example generated from the PrOntoQA-OOD dataset. The examples form the first part of the input prompt. Each example may be different across different test questions resulting in different CoT prompts in each question. 

For each exemplar, the text of the premises is appended after "Q: ", followed by ". Prove: ". The text of the conclusion is appended after ". Prove: ", followed by "\textbackslash nA: ". The gold CoT text is then appended after "\textbackslash nA: ". Two next-line symbols are inserted between two exemplars and between the last exemplar and the text with "Q: " concatenated with the text of the premises from the test question, ". Prove: ", the text of the conclusion from the test question, and "\textbackslash nA: ".

The input prompt consists of 8 examples for the models from GPT2 \citep{radford2019language}, SmolLM2 \citep{allal2024SmolLM2}, OpenELM \citep{mehta2024openelm}, TinyLlama \citep{zhang2024tinyllama}, Stable LM 2 \citep{bellagente2024stable}, and Gemma 2 \citep{team2024gemma}, and 3 examples for the models from GPT2 \citep{radford2019language}. The reason that the models from GPT2 \citep{radford2019language} use fewer examples is that models from GPT2 \citep{radford2019language} can only use a maximum of 1024 input tokens which is not enough for the CoT test using the PrOntoQA-OOD dataset \citep{saparov2023testing} with proof by contradiction. The details of the exemplars in the six deductive rules from the PrOntoQA-OOD dataset \citep{saparov2023testing} are described in Appendix~\ref{PrOntoQA_ex_detals}

\subsubsection{String parsing strategy for the PrOntoQA-OOD dataset}


For the CoT experiment on the PrOntoQA-OOD dataset \citep{saparov2023testing}, given the CoT prompt $x_{CoT}$, the output of the LM is a string which can be represented as $f_\theta(x_{CoT}) = [x_{CoT}, \mathcal{T}_{response}, y_{else}]$, where $\mathcal{T}_{response}$ is the chain-of-thought generated by the LM, and $y_{else}$ is a string that contains content irrelevant to the test question. 

In our string parsing process, we aim to extract the chain-of-thought, $\mathcal{T}_{response}$, generated by the LM. We first remove the CoT prompt, $x_{CoT}$, from $f_\theta(x_{CoT})$ and then extract the string before the first "Q:".

\subsection{Language models}

\subsubsection{GPT-2}

We used the models from GPT-2 \citep{radford2019language} with 117, 345, 762, and 1542 million parameters and noted them as the gpt2, gpt2-medium, gpt2-large, and gpt-xl model, respectively, the same as their names in the Hugging face library. The gpt2, gpt2-medium, gpt2-large, and gpt-xl model are decoder-only transformers based on \citep{radford2018improving} with 12, 24, 36, and 48 layers and a dimension of 768, 1024, 1280, and 1600 in the outputs of all the sub-layers and the embedding layers, respectively. All of the four models used 50,257 vocabularies, a context size of 1024 tokens, and trained on their designed dataset, WebText \citep{radford2019language}, which contains 45 million links of web pages with over 8 million documents and 40 GB of text.

\subsubsection{SmolLM2}

We used the models from SmolLM2 \citep{allal2024SmolLM2} with 135, 360, and 1700 million parameters and noted them as the SmolLM1-135M, SmolLM1-360M, and SmolLM1-1.7B model, respectively, and used the instruction tuned version of these models noted as the SmolLM1-135M-Instruct, SmolLM1-360M-Instruct, and SmolLM1-1.7B-Instruct model, respectively. The models are decoder-only transformers incorporating Grouped-Query Attention (GQA) \citep{ainslie-etal-2023-gqa} and trained on Python-Edu \citep{allal2024SmolLM2, li2023starcodersourceyou}, which contains educational Python samples from The Stack, FineWeb-Edu \citep{penedo2024the}, which has educational web samples from FineWeb, DCLM, and their curated datasets. SmolLM1-1.7B-Instruct is fine-tuned on the SmolTalk dataset \citep{allal2024SmolLM2} and the SmolLM1-135M-Instruct and the SmolLM1-360M-Instruct models are fine-tuned on a subset of the SmolTalk dataset. The instructed models are applied Direct Preference Optimization (DPO) \citep{NEURIPS2023_a85b405e} on the pre-processed version of the UltraFeedback dataset \citep{cui2024ultrafeedbackboostinglanguagemodels}.


\subsubsection{OpenELM}

We used the models from OpenELM \citep{mehta2024openelm} with 270, 450, 1100, and 3000 million parameters and noted them as the OpenELM-270M, OpenELM-360M, OpenELM-1\_1B, and OpenELM-3B model, respectively, and used the instruction tuned version of these models noted as the OpenELM-270M-Instruct, OpenELM-360M-Instruct, OpenELM-1\_1B-Instruct, and OpenELM-3B-Instruct model, respectively. These models adopt decoder-only transformer-based architecture and incorporate GQA \citep{ainslie-etal-2023-gqa}, flash attention \citep{NEURIPS2022_67d57c32}, pre-normalization using RMSNorm \citep{NEURIPS2019_1e8a1942}, rotary positional embedding \citep{su2024roformer}, SwiGLU feed forward network \citep{shazeer2020gluvariantsimprovetransformer}, and the tokenizer from LLama \citep{touvron2023llama2openfoundation}. The models with 270, 450, 1100, and 3000 million parameters have 16, 20, 28, and 36 layers, 1280, 1536, 2048, and 3072 embedding dimensions, and a head dimension of 64, 64, 64, and 128, respectively. All the models have a context length of 2048 and are trained on RefinedWeb \citep{penedo2023refinedwebdatasetfalconllm}, the Github, 
Books, ArXiv, Wikipedia, StackExchange, and C4 subsets in RedPajama \citep{weber2024redpajama}, PILE \citep{gao2020pile800gbdatasetdiverse}, and The Stack, Reddit, PeS2o, Project 
Gutenberg, and Wikipedia + Wikibooks subsets in Dolma \citep{soldaini2024dolmaopencorpustrillion}. The OpenELM-270M-Instruct, OpenELM-360M-Instruct, OpenELM-1\_1B-Instruct, and OpenELM-3B-Instruct model are instruction tuned on the UltraFeedback dataset \citep{cui2024ultrafeedbackboostinglanguagemodels}.

\subsubsection{TinyLlama}

We used the decoder-only Transformer model with 1.1 billion parameters from \citet{zhang2024tinyllama} and noted it as the TinyLlama\_v1\_1 model. The TinyLlama\_v1\_1 model has a context length of 2048, 32 heads, 22 layers, and 32000 vocabularies. The TinyLlama\_v1\_1 model incorporates GQA \citep{ainslie-etal-2023-gqa}, flash attention \citep{NEURIPS2022_67d57c32}, pre-normalization using RMSNorm \citep{NEURIPS2019_1e8a1942}, rotary positional embedding \citep{su2024roformer}, and SwiGLU feed forward network \citep{shazeer2020gluvariantsimprovetransformer} similar to \citet{mehta2024openelm}. The training datasets of the TinyLlama\_v1\_1 model include SlimPajama \citep{cerebras2023slimpajama} derived from RedPajama \citep{weber2024redpajama} and the StarCoder Training Dataset used to train StarCoder \cite{li2023starcoder}.

\subsubsection{Stable LM 2}

We used two decoder-only Transformer models from Stable LM 2 \citep{bellagente2024stable}. One is a 1.6 billion parameter decoder-only language model noted as the stablelm-2-1\_6b model and another one is the fine-tuned version of the stablelm-2-1\_6b model, noted as the
stablelm-2-zephyr-1\_6b model, with DPO \citep{NEURIPS2023_a85b405e} following the Zephyr training recipe \citep{tunstall2023zephyrdirectdistillationlm}. The stablelm-2-1\_6b model contains 1,644,417,024 parameters, 24 layers, 32 heads, and 4096 context length. The stablelm-2-1\_6b model uses rotary positional embedding \citep{su2024roformer} on the first 25\% of head embedding dimensions and LayerNorm \citep{ba2016layernormalization} with learned bias terms as opposed to RMSNorm \citep{zhang2019rootmeansquarelayer}. Only the key, query, and value projection contain bias terms \citep{bai2023qwentechnicalreport}. The training data for the stablelm-2-1\_6b model includes RefinedWeb \citep{penedo2023refinedwebdatasetfalconllm}, RedPajama \citep{weber2024redpajama}, subsets of 
PILE \citep{gao2020pile800gbdatasetdiverse}, and The Stack \cite{li2023starcoder}, OpenWebText \citep{Gokaslan2019OpenWeb}, OpenWebMath \citep{paster2024openwebmath}, OSCAR subset in CulturaX \citep{nguyen-etal-2024-culturax}, 
FanFics, and the Restruct-v1 \citep{bellagente2024stable}. The data for supervised fine-tuning of stablelm-2-zephyr-1\_6b model includes ultrachat\_200k \citep{ding-etal-2023-enhancing}, MetaMathQA \citep{yu2024metamath}, 
WizardLM\_evol\_instruct\_V2\_196k \citep{xu2024wizardlm}, SlimOrca \citep{SlimOrca}, 
openchat\_sharegpt4\_dataset \citep{chen2025sharegpt4v}, Capybara \citep{daniele2023amplify-instruct}, and deita-10k-v0 \citep{liu2024what}. The data for DPO of stablelm-2-zephyr-1\_6b model includes UltraFeedback dataset \citep{cui2024ultrafeedbackboostinglanguagemodels} and Intel Orca Pairs.

\subsubsection{Gemma 2}

We used two decoder-only Transformer models from Gemma 2 \citep{team2024gemma}: Gemma 2 IT 2B and Gemma 2 IT 9B, noted as gemma-2-2b-it and gemma-2-9b-it. The gemma-2-2b-it model has 2,614,636,800 model parameters, 2304 embedding dimensions, 26 layers, and 8 heads. The gemma-2-9b-it model has 9,242,164,736 model parameters, 42 layers, and 16 heads. Both the gemma-2-2b-it and gemma-2-9b-it models have a context length of 8192 tokens and 256128 vocabularies. The architecture of the models from Gemma 2 \citep{team2024gemma} are based on the architecture from Gemma 1 \citep{gemmateam2024gemmaopenmodelsbased} and incorporates rotary positional embedding \citep{su2024roformer}, approximated GeGLU non-linearity, GQA \citep{ainslie-etal-2023-gqa}, RMSNorm \citep{zhang2019rootmeansquarelayer}, their proposed Logit soft-capping, and local sliding window attention \citep{beltagy2020longformerlongdocumenttransformer} in some of the layers and global sliding window attention \citep{luong-etal-2015-effective} in the other layers. The training data includes web documents, code, and science articles, and the link, name, or citation of the data source is not included in their paper \citep{team2024gemma}. 

We loaded the two models with bfloat16 in the CoT experiment on the CSQA dataset \cite{talmor-etal-2019-commonsenseqa} and the PrOntoQA-OOD dataset \citep{saparov2023language}. We loaded the gemma-2-9b-it model with float16 in the attention analysis in order to extract the attention weights.

\subsection{Fine-tuning}
\label{finetune_method}

To enhance the reasoning capabilities of models below the identified parameter threshold ($\sim$1.6 billion), we fine-tuned six of the smallest models-gpt2, SmolLM2-135M, SmolLM2-135M-Instruct, OpenELM-270M, OpenELM-270M-Instruct, and gpt2-medium models-using task-specific exemplars from the PrOntoQA-OOD dataset \citep{saparov2023testing}. These exemplars are structured logical reasoning examples covering six distinct deductive rules: implication elimination, conjunction introduction, conjunction elimination, disjunction introduction, disjunction elimination, and proof by contradiction. Each fine-tuning instance consisted of 3 exemplars per question, with a total of 100 questions per deductive rule, resulting in 1,800 exemplars in total.

We restricted our selection to models with $\leq$355 million parameters to ensure feasibility within a single 40 GB GPU during fine-tuning. The dataset was randomly shuffled and split into 90\% training data (1,620 exemplars) and 10\% validation data (180 exemplars). Training data were concatenated and segmented into multiple 1024-token blocks to align with input constraints.

For fine-tuning, we used the Hugging Face Trainer API from the Transformers Python library. The training was framed as causal language modeling, where the input tokens were shifted by one token to serve as the target labels. We employed the Adam optimizer with a weight decay of 0.01, hyperparameters $\beta_1=0.9$, $\beta_2=0.999$, and $\epsilon=10^{-8}$ \citep{loshchilov2019decoupledweightdecayregularization}. Training was conducted for 100 epochs with a batch size of 1,000 and a learning rate of $2\cdot10^{-5}$.

To select the most effective fine-tuned model, we evaluated validation loss across all 100 epochs and chose the model with the best validation loss to avoid overfitting. We refer to these fine-tuned models as gpt2-best, SmolLM2-135M-best, SmolLM2-135M-Instruct-best, OpenELM-270M-best, OpenELM-270M-Instruct-best, and gpt2-medium-best models.

\subsection{Token generation strategy}

We used the "generate" function from the transformers library from Hugging Face. In the CoT experiment using the CSQA dataset \cite{talmor-etal-2019-commonsenseqa}, we set the maximum number of the generated tokens as the number of the input tokens added with 100, set "do\_sample" to be False, and set "num\_beams" to be 1 such that each token is sampled from the output of the transformer using greedy decoding. In the CoT experiment using the PrOntoQA-OOD dataset \citep{saparov2023testing}, we used greedy decoding by making "do\_sample" to be False and set "num\_beams" to be 1. We set the maximum number of the generated tokens as 256 which is the same as the number provided in the "opt.py" sample code in the PrOntoQA-OOD dataset \citep{saparov2023testing}. We set the repetition penalty to $0.0001$ for both of the experiments to prevent the LM from producing the output with empty text.

\section{Results}

\subsection{Commonsense Reasoning in Multiple-Choice Question Answering}

\begin{figure}[!t]
\begin{center}
\centerline{\includegraphics[width=\columnwidth]{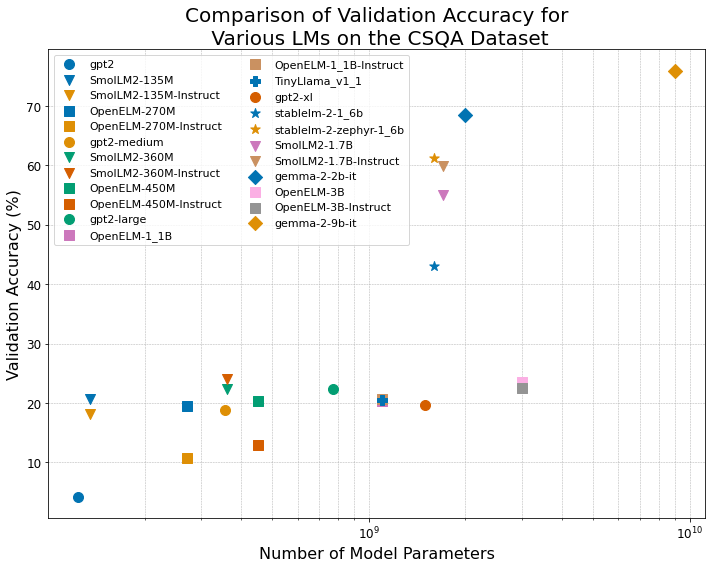}}
\caption{Accuracy plot of different LMs solving 1221 multiple-choice questions from the validation set in the CSQA dataset \citep{talmor-etal-2019-commonsenseqa}. The circle markers represent different models from GPT2 \citep{radford2019language}, the upside-down triangle markers represent different models from SmolLM2 \citep{allal2024SmolLM2}, the square markers represent different models from OpenELM \citep{mehta2024openelm}, the plus marker represents the 1.1B model from TinyLlama \citep{zhang2024tinyllama}, the star markers represent different models from Stable LM 2 \citep{bellagente2024stable}, and the diamond markers represent different models from Gemma 2 \citep{team2024gemma}. Different colors are used to differentiate different models from the same family of models.}
\label{Figure_CSQA_acc_sample_false}
\end{center}
\end{figure}

We evaluate the commonsense reasoning capabilities of 23 different LMs using the CSQA dataset \citep{talmor-etal-2019-commonsenseqa}. Model accuracy is computed as the number of correct answers divided by the total number of questions. Our analysis reveals a clear trend of increasing accuracy as model size increases, with a significant performance jump between models with fewer than 1.5 billion parameters and those exceeding 1.6 billion parameters, except for models from OpenELM \citep{mehta2024openelm}. The gpt2 model \citep{radford2019language} achieves the lowest accuracy of 4.18\%, whereas the gemma2-9b-it model \citep{team2024gemma} reaches the highest accuracy of 75.92\% as shown in Figure~\ref{Figure_CSQA_acc_sample_false}.

\begin{figure}[ht]
\begin{center}
\centerline{\includegraphics[width=\columnwidth]{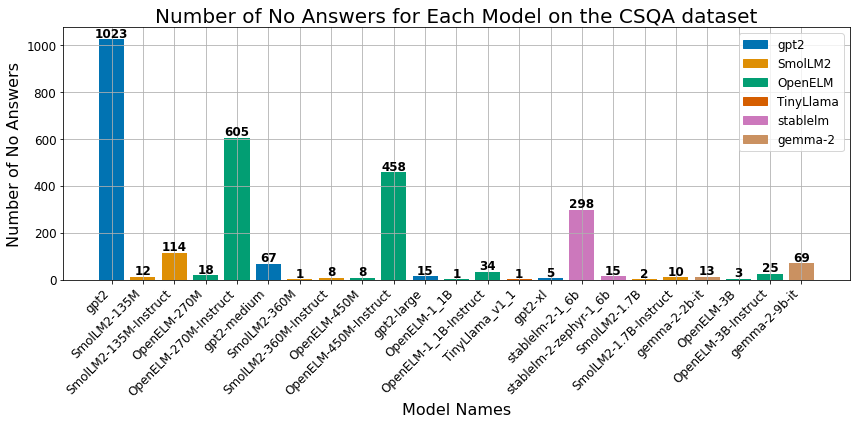}}
\caption{Number of responses that cannot be correctly parsed to get the answer from different LMs solving 1221 multiple-choice questions from the validation set in the CSQA dataset \citep{talmor-etal-2019-commonsenseqa}. The blue, orange, green, red, pink, and brown bars show the counts obtained from the models in GPT2 \citep{radford2019language}, SmolLM2 \citep{allal2024SmolLM2}, OpenELM \citep{mehta2024openelm}, TinyLlama \citep{zhang2024tinyllama}, Stable LM 2 \citep{bellagente2024stable}, and Gemma 2 \citep{team2024gemma}, respectively.}
\label{Figure_CSQA_no_answer_sample_false}
\end{center}
\end{figure}

A noticeable accuracy increase is observed between gpt2-xl (1.542B parameters) \citep{radford2019language} with an accuracy of 19.57\% and stablelm-2-1\_6B \citep{bellagente2024stable} with 43.00\%, supporting the hypothesis that reasoning abilities improve significantly beyond a certain parameter threshold as shown in Figure~\ref{Figure_CSQA_acc_sample_false}. Additionally, we analyze the number of unparseable responses using the method described in Section~\ref{String_parse_CSQA} as shown in Figure~\ref{Figure_CSQA_no_answer_sample_false}. gpt2, OpenELM-270M-Instruct, OpenELM-450M-Instruct, and stablelm-2-1\_6B exhibit a high failure count of over 200, indicating that sub-threshold models struggle with structured answer formatting.


\subsection{Deductive Reasoning in Language Models}

To assess deductive reasoning capabilities, we evaluate models on six logical reasoning tasks from the PrOntoQA-OOD dataset \citep{saparov2023testing}. The correctness of the generated CoT is assessed by comparing the generated steps to the gold CoT using the "analyze\_results.py" evaluation script from \citet{saparov2023testing}.

\begin{figure*}[ht]
\begin{center}
\centerline{\includegraphics[width=\textwidth]{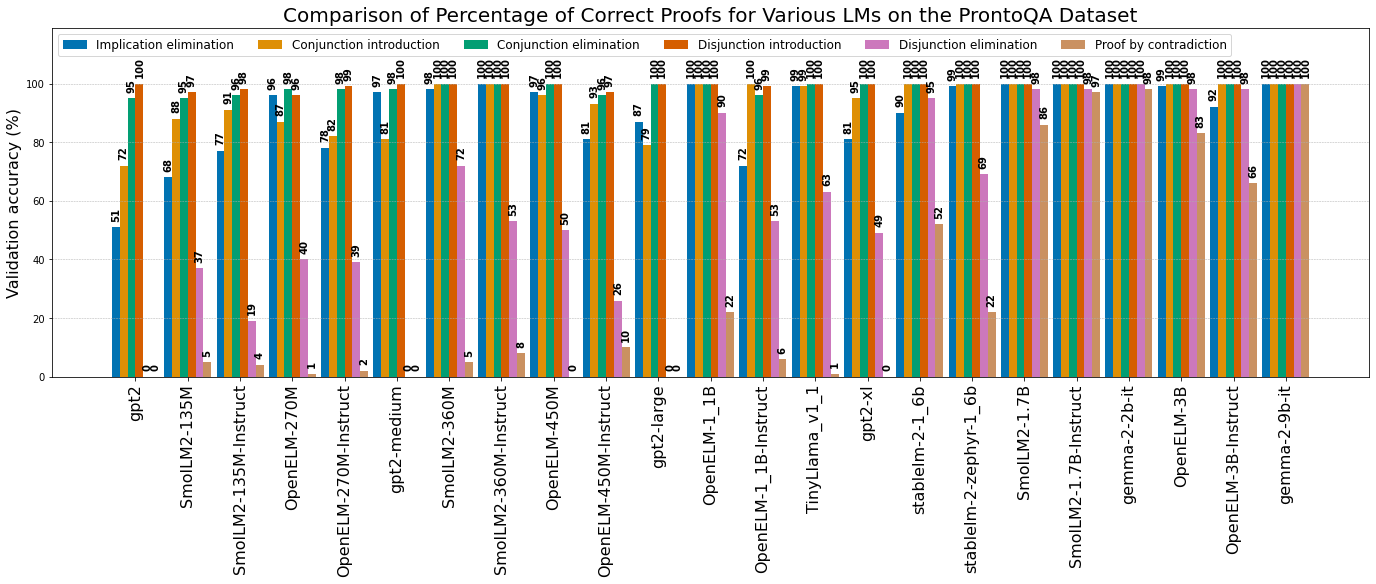}}
\caption{Accuracy of different LMs solving 100 deductive reasoning questions generated by the PrOntoQA-OOD data generation codes \citep{saparov2023testing} on six different deduction rules for each model.}
\label{Figure_ProntoQA}
\end{center}
\end{figure*}

A distinct parameter gap is observed in multiple reasoning tasks as shown in Figure~\ref{Figure_ProntoQA} and in Figure~\ref{Figure_ProntoQA_scatter}:

\begin{itemize}
    \item Implication Elimination: Accuracy increases from 68\% in SmolLM2-135M (135M parameters) to 78\% in OpenELM-270M-Instruct (270M parameters) as shown in Figure~\ref{fig:scatter_IE}.
    \item Disjunction Elimination: A gap is observed between gpt2-large (762M, 0\% accuracy) and OpenELM-1.1B-Instruct (1.1B, 53\% accuracy), suggesting a threshold around 1.1 billion parameters as shown in Figure~\ref{fig:scatter_DE}.
    \item Proof by Contradiction: Models with fewer than 1.5 billion parameters, including gpt2-xl (1.542B) with 0\% accuracy, fail to generate correct proofs, whereas stablelm-2-zephyr-1\_6b achieves 22\% accuracy as shown in as shown in Figure~\ref{fig:scatter_PBC}.
\end{itemize}

For simpler reasoning tasks, such as conjunction introduction and conjunction elimination, all models achieve above 95\% accuracy, indicating that smaller models can effectively handle lower-complexity deductions as shown in Figure~\ref{fig:scatter_CI}, Figure~\ref{fig:scatter_CE}, and Figure~\ref{fig:scatter_DI}.

\subsection{Fine-Tuning Improves Sub-Threshold Models}

We evaluate whether fine-tuning enhances reasoning performance in sub-threshold models by training six of the smallest models-gpt2, SmolLM2-135M, SmolLM2-135M-Instruct, OpenELM-270M, OpenELM-270M-Instruct, and gpt2-medium-on exemplars from the PrOntoQA-OOD dataset \citep{saparov2023testing}. The fine-tuned models exhibit substantial improvements in implication elimination, conjunction introduction, conjunction elimination, and disjunction introduction, all achieving accuracy above 90\% as shown in Figure~\ref{Figure_ProntoQA_finetuned}. However, performance remains inconsistent for more complex reasoning tasks. While fine-tuning benefits certain models, others experience a drop in performance:

\begin{itemize}
    \item Proof by Contradiction: Fine-tuning increases accuracy in SmolLM2-135M-Instruct (3\% to 42\%) and OpenELM-270M (3\% to 46\%). However, SmolLM2-135M drops from 3\% to 0\%, and OpenELM-270M-Instruct experiences a slight change from 11\% to 12\% as shown in Figure~\ref{Figure_ProntoQA_finetuned}.
    \item Disjunction Elimination: Fine-tuning fails to improve accuracy for gpt2 and gpt2-medium (both remain at 0\%), while SmolLM2-135M drops significantly from 19\% to 0\%, and SmolLM2-135M-Instruct decreases from 13\% to 0\%. In contrast, the OpenELM-270M model improves from 41\% to 66\%, though OpenELM-270M-Instruct also experiences a performance decline from 30\% to 4\% as shown in Figure~\ref{Figure_ProntoQA_finetuned}.
\end{itemize}

These results indicate that fine-tuning is effective for shorter reasoning chains but struggles with longer proofs requiring disjunction elimination or proof by contradiction.

\subsection{Attention Map Analysis}

In this section, we analyze the differences in attention maps between LMs capable of generating correct CoT reasoning and those that fail. We compare the attention maps of the gpt2 model and the gemma2-9b-it model (loaded with float16) on the first proof of the implication elimination task from the PrOntoQA-OOD dataset \citep{saparov2023testing}. We construct a prompt consisting of three exemplars with implication elimination generated from the PrOntoQA-OOD dataset \citep{saparov2023testing}.
The corresponding test question is: "Q: Sterpuses are transparent. Wren is a sterpus. Prove: Wren is transparent."
The expected gold CoT is: "Wren is a sterpus. Sterpuses are transparent. Wren is transparent."
The gemma2-9b-it model correctly generates the gold CoT. However, the gpt2 model, after applying our string parsing method, produces: "Wren is a sterpus. Sterpuses are transparent. Wren is a sterpus."
The final phrase, "a sterpus", is incorrect-gpt2 repeats part of the premise instead of concluding with "transparent." This discrepancy shows a failure in reasoning.

\begin{figure}[!t]
    \centering
    \subfigure[The gpt2 model]{\includegraphics[width=0.99\columnwidth]{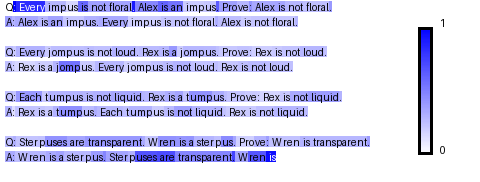}\label{fig:toekn_score_gpt2}}
    \subfigure[The gemma2-9b-it model]{\includegraphics[width=0.99\columnwidth]{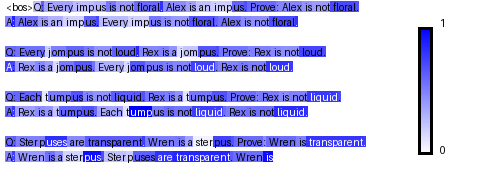}\label{fig:toekn_score_gemma2-9b-it}}
    \caption{Visualization of the normalized token-level scores \citep{kang-shin-2023-samrank} using the first head in the last layer from the gpt2 model and the gemma2-9b-it model loaded with float16. Both of the models are prompted with the CoT prompt corresponding to the first proof of the implication elimination task from the PrOntoQA-OOD dataset \citep{saparov2023testing} concatenated with "Wren is a sterpus. Sterpuses are transparent. Wren is" The saturation of the background color is proportional to the normalized token-level scores \citep{kang-shin-2023-samrank} as indicated in the color bar on the right of each figure.}
\label{Figure_ProntoQA_toekn_score}
\end{figure}

To analyze why gemma2-9b-it succeeds while gpt2 fails, we compute token-level scores from the attention maps using the method described in \citet{kang-shin-2023-samrank}. Given an attention matrix $A \in \mathbb{R}^{n \times n}$, the global attention score $G_{t_i}$ for each token $i$ is computed as, $G_{t_i} = \sum_{j=1}^{n} A_{ji}$. To focus on token-to-token interactions beyond the first token, we set $G_{t_i} = 0$, as suggested in \citet{kang-shin-2023-samrank}. We then compute the proportional attention score, $P_{t_i} = \sum_{j=1}^{n} B'_{ij}$, where $B'$ is derived from $B'_{ji} = \frac{B_{ji}}{\sum_{j=1}^{n} B_{ji}}$ and  $\quad B = A \cdot \text{diag}(G)$. Finally, the token-level scores are given by $S_{t_i} = G_{t_i} + P_{t_i}$. These scores are then normalized using $S'_{t_i} = \frac{S_{t_i} - \text{min}(S_{t_i})}{\text{max}(S_{t_i}) - \text{min}(S_{t_i})}$ and visualized with a color-coded background proportional to $S'_{t_i}$ as shown in Figure~\ref{Figure_ProntoQA_toekn_score}.

The visualization of gpt2's attention scores of the first head in the last layer in Figure~\ref{fig:toekn_score_gpt2} reveals that the model assigns high attention to irrelevant tokens such as "Every" and "is". In contrast, gemma2-9b-it assigns higher attention scores of the first head in the last layer to tokens such as "loud," "liquid," and most importantly, "transparent", as shown in Figure~\ref{fig:toekn_score_gemma2-9b-it}.

The results imply that models capable of generating correct CoTs tend to assign higher token-level attention scores to:
\begin{itemize}
    \item The next expected token in the reasoning process.
    \item Tokens that correspond to the correct parts of speech (e.g., adjectives following a "be" verb).
\end{itemize}

For models that fail to generate correct CoTs (e.g., gpt2), token-level scores for the correct next token or its relevant grammatical components remain low. This indicates that attention allocation may play a crucial role in logical inference and CoT generation.

The differences in attention between the two models are further studied in a detailed per-head comparison. We observe that, except for the first key embedding, the attention scores between the query embedding of the first head in the Gemma2-9B-IT model and the key embedding of the "transparent" token—the correct token to be generated—are higher compared to other attention scores in the first, second, third, fifth, sixth, seventh, eighth, ninth, tenth, eleventh, twelfth, thirteenth, and fourteenth attention heads, as shown in Figure~\ref{Figure_ProntoQA_MP_attention_gemma2-9b-it}.

However, in the GPT-2 model, except for the first key embedding, the attention scores between the query embedding of the first head and the key embedding of the "ster" token—the incorrect token to be generated—are higher compared to other attention scores in the second, third, tenth, and eleventh attention heads, as shown in Figure~\ref{Figure_ProntoQA_MP_attention_gpt2}.

\section{Conclusion}

We investigate the commonsense and deductive reasoning abilities of 23 selected LMs through CoT prompting \citep{wei2022chain}. Our findings reveal a critical parameter threshold ($\sim$1.6 billion) beyond which reasoning abilities significantly improve, as demonstrated by success rates in the CommonsenseQA dataset \citep{talmor-etal-2019-commonsenseqa} and deductive reasoning tasks in the PrOntoQA-OOD dataset \citep{saparov2023language}. In particular, we observe a parameter gap for proof by contradiction between models with fewer than 1.5 billion and more than 1.6 billion parameters and a smaller gap (774 million to 1.1 billion parameters) for disjunction elimination.

To address limitations in sub-threshold models, we fine-tune six of the smallest LMs-GPT2, SmolLM2-135M, SmolLM2-135M-Instruct, OpenELM-270M, OpenELM-270M-Instruct, and GPT2-Medium—using task-specific exemplars from the PrOntoQA-OOD dataset \citep{saparov2023language}. Fine-tuning substantially improves reasoning performance, enabling these models to generate accurate CoTs without exemplars in the prompt for four deductive rules—implication elimination, conjunction introduction, conjunction elimination, and disjunction introduction. However, tasks requiring longer reasoning chains, such as proof by contradiction and disjunction elimination, remain challenging for fine-tuned models with limited parameters.

Finally, our attention map analysis provides insights into the interpretability of reasoning processes. We show that models capable of generating correct CoTs, such as gemma2-9b-it, exhibit higher token-level scores \citep{kang-shin-2023-samrank} for the next correct token and for tokens corresponding to the correct parts of speech, whereas smaller models, such as gpt2, fail to capture these patterns. These findings contribute to a deeper understanding of reasoning in decoder-only transformer-based models, offering practical implications for model selection, fine-tuning strategies, and interpretability in language model research.




\section*{Impact Statement}

Understanding how language models develop reasoning abilities is essential for improving their performance in complex decision-making tasks. This work identifies a critical parameter threshold ($\sim$1.6 billion) beyond which reasoning abilities significantly improve, providing a guideline for selecting models suited for reasoning tasks. We also show that fine-tuning smaller models can help them achieve reasoning performance closer to that of larger models, making them more effective while using fewer computational resources. Additionally, our analysis of attention maps reveals how models allocate focus during reasoning, offering a valuable tool for improving interpretability and debugging. These findings contribute to better model design, optimization strategies, and transparency in AI reasoning systems.




\bibliography{example_paper}
\bibliographystyle{icml2025}

\newpage
\appendix
\onecolumn
\section{Appendix}
\subsection{Details of the exemplars in the six deductive rules from the PrOntoQA dataset}
\label{PrOntoQA_ex_detals}

\subsubsection{Details of the exemplars of the implication elimination}

The example with implication elimination contains two premises and one conclusion. One premise is a sentence that shows an implication of an antecedent and a consequent; another premise is a declarative sentence called propositional statement that can be answered either true or false. The conclusion is a propositional statement to be answered through the CoT reasoning by implication elimination. 
The gold CoT text will first repeat the propositional statement, then repeat the implication sentence, and finally repeat the propositional statement in the conclusion. All of the sentences are a type of copular sentence structure, which consists of a subject, followed by a 'be' verb, which links the subject to a complement \citep{moro2006copular}. For instance, the premise in one of the examples is "Every impus is not floral. Alex is an impus.", in which "Every impus is not floral" is a premise with an implication sentence, "Alex is an impus" is a premise with a propositional statement, and "Alex is not floral", the conclusion of the example is "Alex is not floral", and the corresponding gold CoT is "Alex is an impus. Every impus is not floral. Alex is not floral."

\subsubsection{Details of the exemplars of the conjunction introduction}

The examples with conjunction introduction contain two premises and one conclusion. Both the premises and the conclusion are propositional statements. All of the sentences are copular sentences with a 'be' verb. One propositional statement contains a fictional noun in the complement and the other an adjective. The two premises are randomly ordered in each example. For instance, in "Alex is earthy. Alex is an impus.", the two premises are "Alex is earthy" and "Alex is an impus", "Alex" is the subject, "earthy" is an adjective, and "impus" is a fictional noun. The conclusion of the preceding is "Alex is an earthy impus". 
The gold CoT text will first repeat the premise with the fictional noun, then repeat the premise with the adjective, and finally repeat the propositional statement in the conclusion. Therefore, the gold CoT of the preceding instance is "Alex is an impus. Alex is earthy. Alex is an earthy impus."

\subsubsection{Details of the exemplars of the conjunction elimination}

The examples with conjunction elimination contain one premise and one conclusion. All of the sentences are copular sentences with a 'be' verb. Both the premise and the conclusion are propositional statements with the same subject. The complement in the premise is an adjective-noun phrase and the complement in the conclusion is a noun in the premise. 
The Gold CoT will first repeat the premise and then repeat the conclusion as: "Max is a hot impus. Max is an impus." when the premise is "Max is a hot impus." and the conclusion is "Max is an impus."

\subsubsection{Details of the exemplars of the disjunction introduction}

The examples with disjunction introduction contain one premise and one conclusion. The premise is a copular sentence with a 'be' verb and a noun as the complement. 
The conclusion is a copular sentence with a 'be' verb and a compound subject complement, which includes a fictional noun and an adjective connected by 'or.' 
The Gold CoT will first repeat the premise and then repeat the conclusion: "Alex is an impus. Alex is earthy or an impus." when the premise is "Alex is an impus. and the conclusion is "Alex is earthy or an impus."

\subsubsection{Details of the exemplars of the disjunction elimination}

The examples with disjunction elimination contain three premises and one conclusion. The first two premises are copular sentences with a 'be' verb and a fictional noun as the complement, which shares the same word in both premises but may appear in either singular or plural form depending on the subject of the sentence. The subject in the first two premises is either a fictional plural countable noun or a noun phrase in which the determiner is either "Every" or "Each" and the noun is a fictional countable noun. The last premise is a copular sentence with a 'be' verb and a compound subject complement, which includes both of the fictional nouns in the subject of the two premises connected by 'or.' 
The conclusion is a copular sentence with a 'be' verb and a complement, which is one of the complements in the first two premises. 
The Gold CoT is composed of three paragraphs. In the first two paragraphs, the first sentence is an imperative sentence starting with "Assume", the second and the third sentences are copular sentences with a 'be' verb. In the first paragraph, the object of the imperative verb 'Assume' is a noun clause, which is a copular sentence with the same subject in the conclusion, a 'be' verb, and the first fictional noun in the compound subject complement of the third premise as the complement. In the second paragraph, the complement of the noun clause is the second fictional noun in the compound subject complement of the third premise. The second sentence in the first two paragraphs is a copular sentence. The second sentence in the first paragraph is the first premise and the second sentence in the second paragraph is the second premise. The third sentence in the two paragraphs is the same as the sentence in the conclusion. The last paragraph contains only one sentence in a conditional structure, with the subordinate clause presenting a reason for the conclusion stated in the main clause. The last paragraph starts with the subordinate clause with "Since" connected with a group of words which is the same as the third premise. The group of words after the subordinate clause is the main clause which is the same as the conclusion.

\subsubsection{Details of the exemplars of the proof by contradiction}

The examples with proof by contradiction contain two premises and one conclusion. The first premise is a complex sentence structure with a main clause and a relative clause. The main clause is a copular sentence with a 'be' verb connecting the subject and the complement, "Everything" as the subject, a noun phrase with a fictional countable noun, and "a" being the article as a complement. The relative clause that modifies the subject in the main clause has 'that' as the subject, 'is' as the copula, and a compound noun phrase as the complement, consisting of two noun phrases—each with an article 'a' and a fictional singular countable noun—connected by the conjunction 'or.' The second premise has a noun as the subject, a 'be' verb, the negation 'not' which negates the verb 'is', and a noun phrase with the article 'a' and a fictional countable noun as the complement. 
The conclusion is a compound sentence with two copular sentences connected by 'and.' Each copular sentence shares the same subject from the second premise, uses 'is' as the copula, includes the negation 'not' which negates the verb 'is,' and contains a noun phrase. The noun phrase in the first copular sentence of the conclusion is to the first noun phrase in the relative clause of the first premise, while the noun phrase in the second copular sentence of the conclusion is to the second noun phrase in the relative clause of the second premise. 
The Gold CoT consists of three paragraphs. In the first two paragraphs, the first sentence is an imperative sentence starting with 'Assume,' in which the object of the imperative verb 'Assume' is a noun clause. This noun clause is a copular sentence. In the first paragraph, the noun clause is the same as the first copular sentence in the conclusion, and in the second paragraph, the noun clause is the same as the second copular sentence in the conclusion. In the first two paragraphs, the second sentence is the same compound sentence as the conclusion but without the negation 'not' in both copular sentences, the third sentence is the same as the first premise, the fourth sentence is the same copular sentence as the second premise but without the negation 'not,' the fifth sentence is a main clause with 'This' as the subject, 'contradicts' as the verb, 'with' as the preposition, and the second premise as the object of the verb 'contradicts,' and the last sentence in the first paragraph is the first copular sentence in the conclusion, while the last sentence in the second paragraph is the second copular sentence in the conclusion. The last paragraph is identical to the conclusion.

\subsection{Figures}

\begin{figure*}[!t]
\centering
\includegraphics[width=\textwidth]{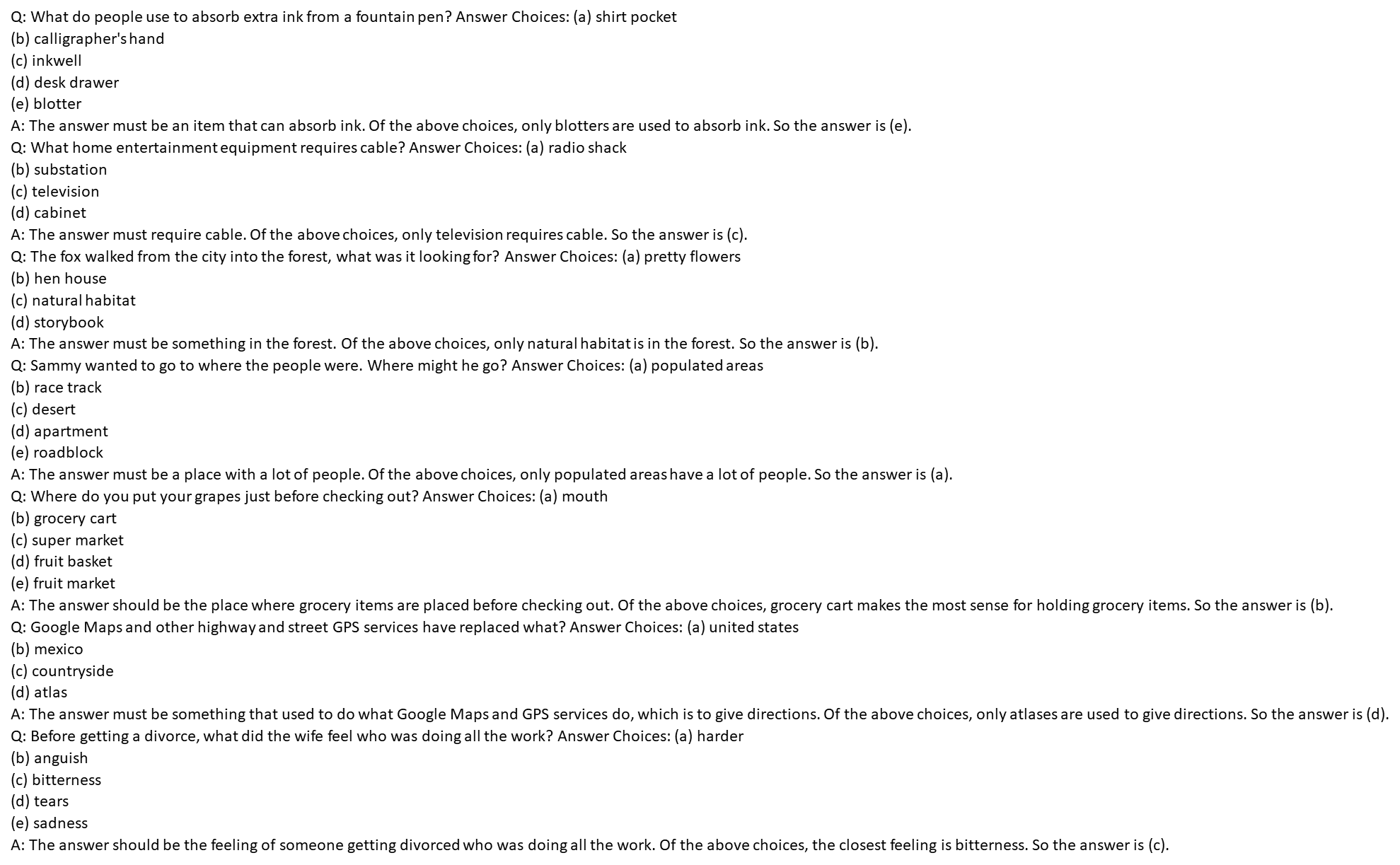}
\caption{The 7 exemplars for the CoT experiments on the CSQA dataset \cite{talmor-etal-2019-commonsenseqa} adopted from \citep{wei2022chain}.}
\label{fig:CSQA_exemplars}
\end{figure*}

\begin{figure*}[!ht]
    \centering
    \subfigure[Implication elimination]{\includegraphics[width=0.42\textwidth]{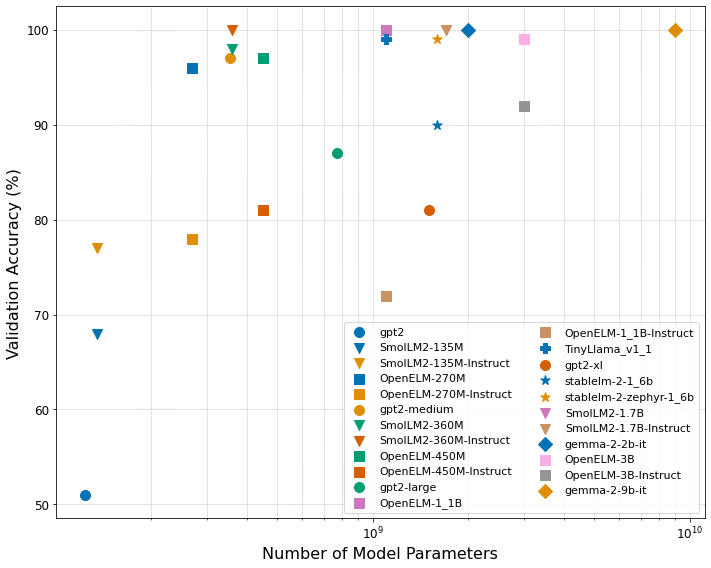}\label{fig:scatter_IE}}
    \subfigure[Conjunction introduction]{\includegraphics[width=0.42\textwidth]{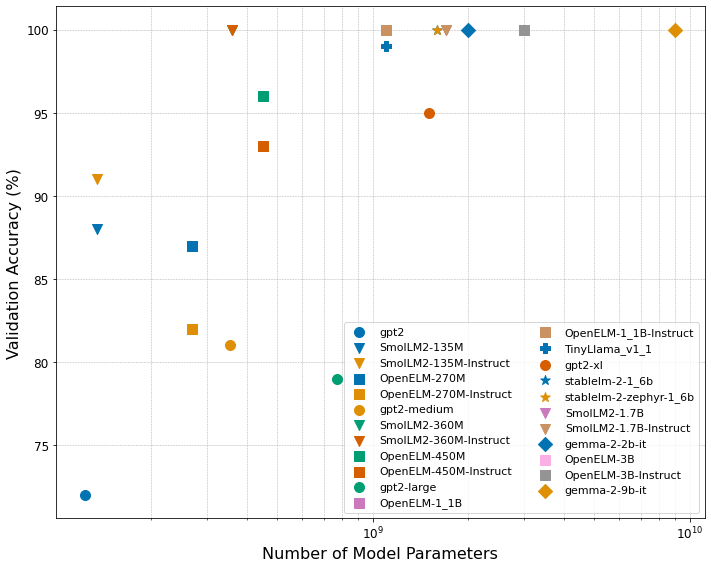}\label{fig:scatter_CI}}
    \subfigure[Conjunction elimination]{\includegraphics[width=0.42\textwidth]{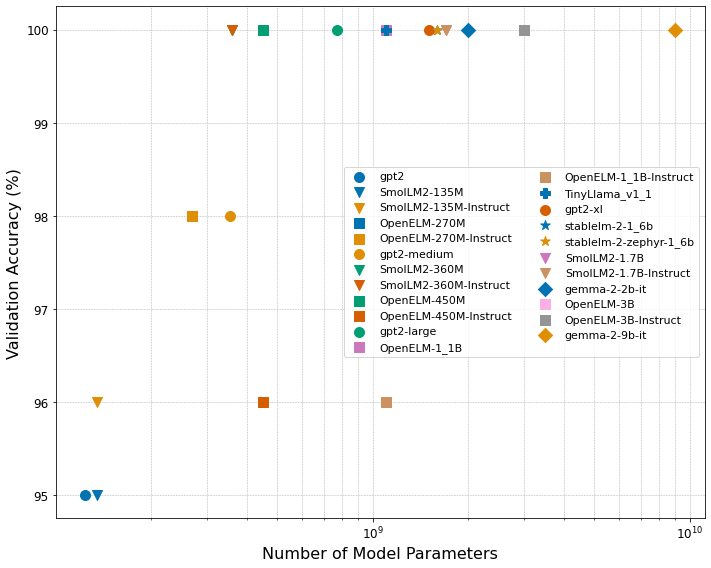}\label{fig:scatter_CE}}
    \subfigure[Disjunction introduction]{\includegraphics[width=0.42\textwidth]{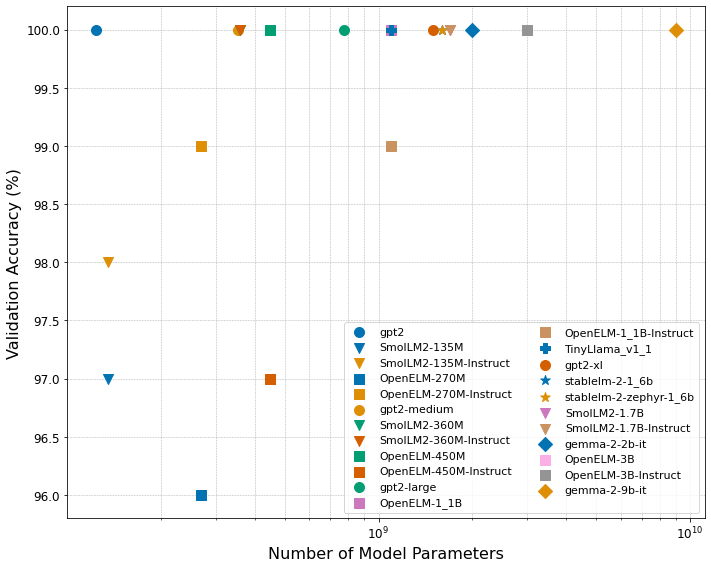}\label{fig:scatter_DI}}
    \subfigure[Disjunction elimination]{\includegraphics[width=0.42\textwidth]{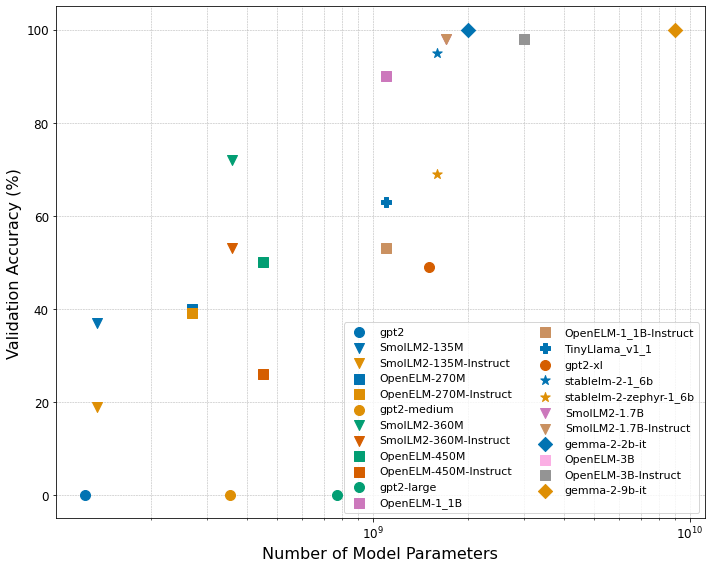}\label{fig:scatter_DE}}
    \subfigure[Proof by contradiction]{\includegraphics[width=0.42\textwidth]{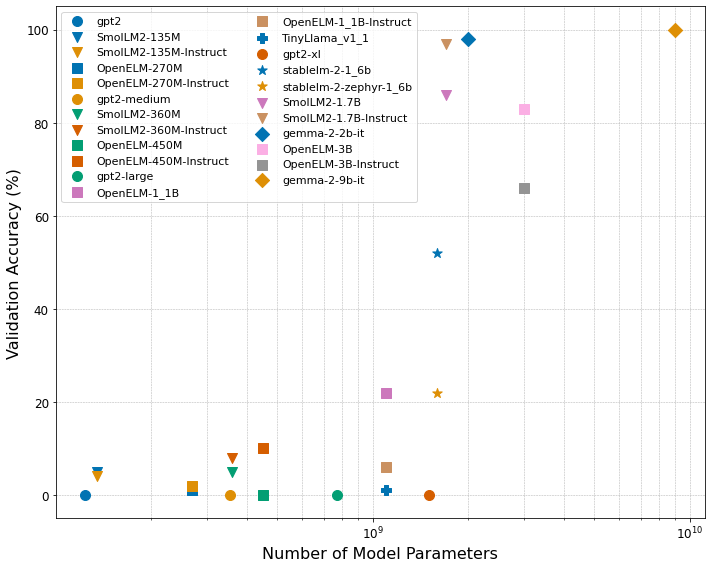}\label{fig:scatter_PBC}}
    \caption{Accuracy plot of different LMs solving 100 deductive reasoning questions generated by the PrOntoQA-OOD data generation codes \citep{saparov2023testing} on six different deduction rules for each model. The circle markers represent different models from GPT2 \citep{radford2019language}, the upside-down triangle markers represent different models from SmolLM2 \citep{allal2024SmolLM2}, the square markers represent different models from OpenELM \citep{mehta2024openelm}, the plus marker represents the 1.1B model from TinyLlama \citep{zhang2024tinyllama}, the star markers represent different models from Stable LM 2 \citep{bellagente2024stable}, and the diamond markers represent different models from Gemma 2 \citep{team2024gemma}. Different colors are used to differentiate different models from the same family of models.}
\label{Figure_ProntoQA_scatter}
\end{figure*}

\begin{figure*}[ht]
\begin{center}
\centerline{\includegraphics[width=\textwidth]{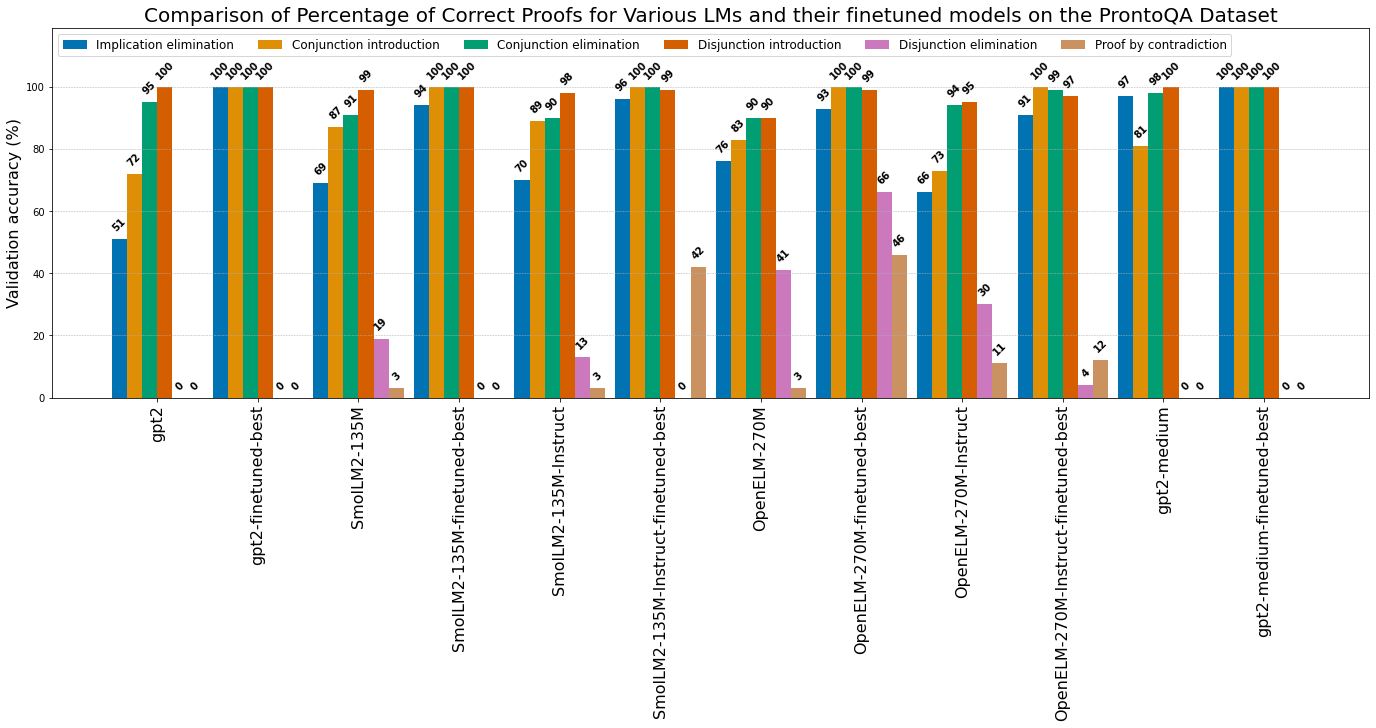}}
\caption{Accuracy of different LMs and the fine-tuned version of these LMs on proving the test questions that are not seen in the exemplars or in the training data. The fine-tuned version of these LMs is trained on 1800 exemplars of 100 deductive reasoning questions for each of the 6 deduction rules generated by the PrOntoQA-OOD data generation codes \citep{saparov2023testing}. The vanilla LMs are prompted with CoT and the fine-tuned version of these LMs are prompted with just the test question.}
\label{Figure_ProntoQA_finetuned}
\end{center}
\end{figure*}


\begin{figure}[!ht]
\begin{center}
\centering
    \setcounter{subfigure}{0}
    \subfigure[The $\text{1}^\text{st}$ head]{\includegraphics[width=0.83\columnwidth]{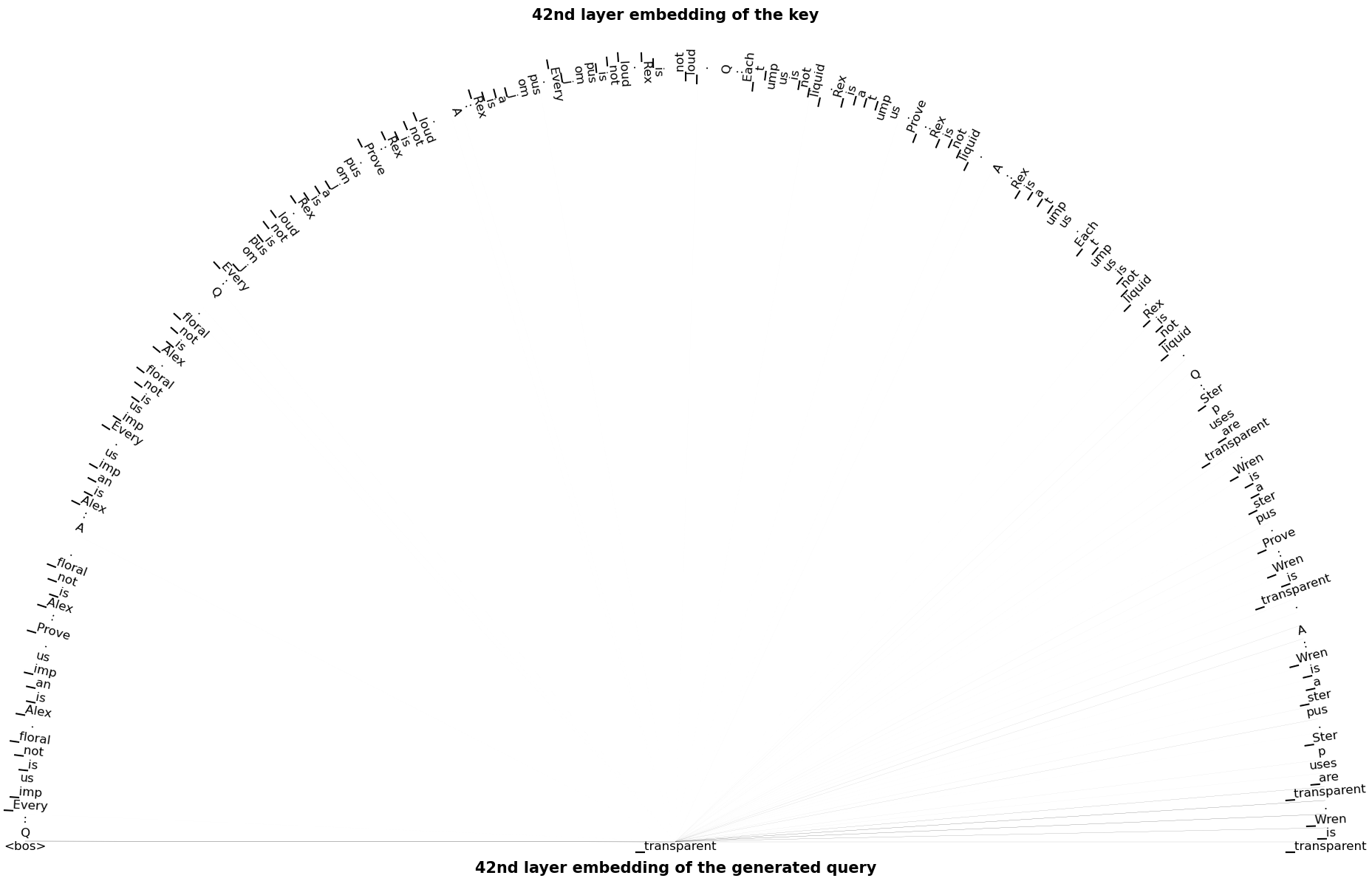}\label{fig:gemma2_H1}}
    \subfigure[The $\text{2}^\text{nd}$ head]{\includegraphics[width=0.83\columnwidth]{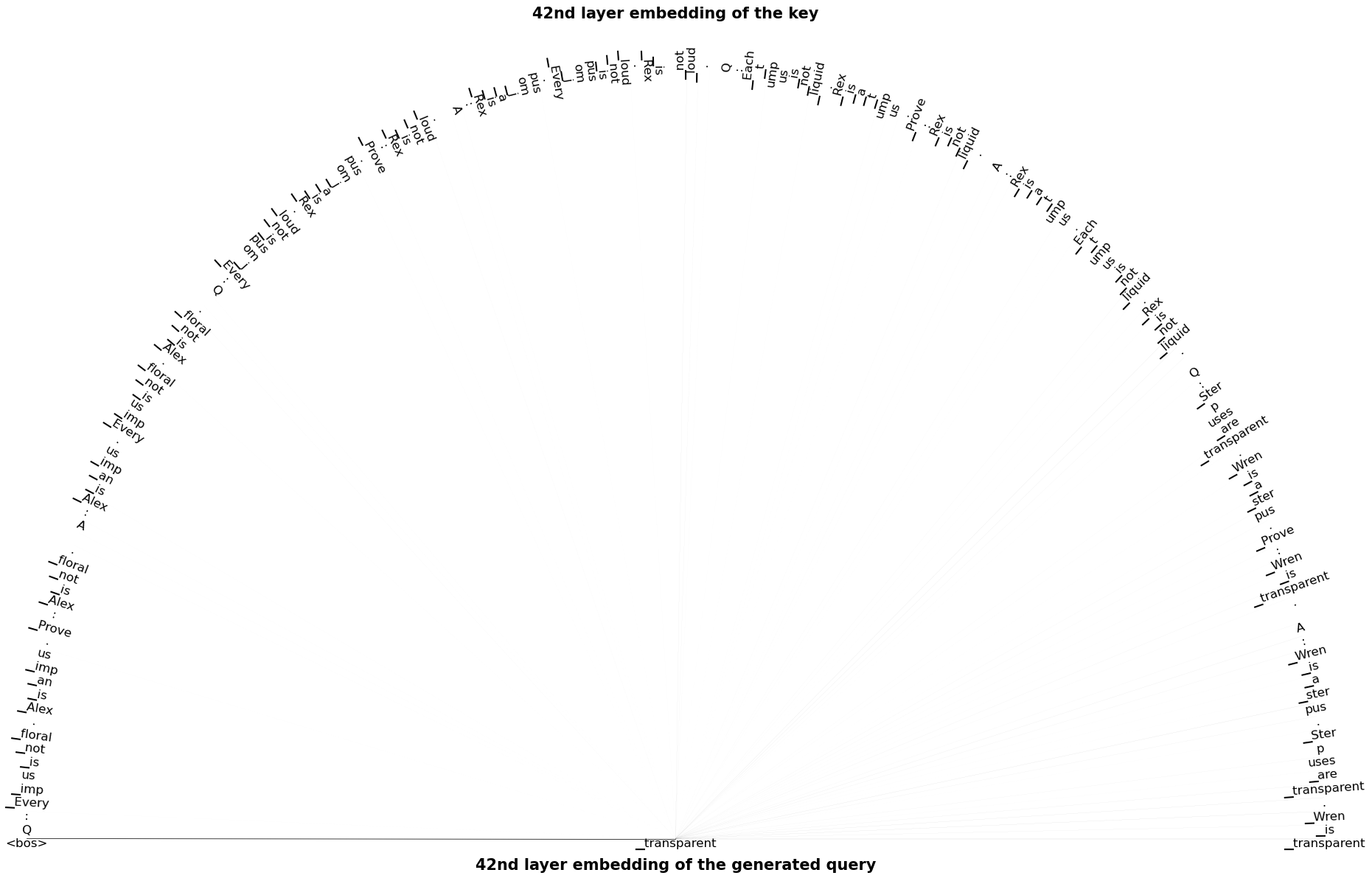}\label{fig:gemma2_H2}}
    \caption{Visualization of attention maps for each head in the final attention layer of the gemma2-9b-it model loaded with float16 precision. The model is prompted using the CoT prompt for the first proof of the implication elimination task from the PrOntoQA-OOD dataset \citep{saparov2023testing}, appended with the text: "Wren is a sterpus. Sterpuses are transparent. Wren is transparent". The thickness of each line connecting token embeddings is proportional to the corresponding attention score. The arc-shaped text represents the key embeddings, progressing from the bottom-left to bottom-right. The text at the bottom represents the query embedding of the last token.}
\label{Figure_ProntoQA_MP_attention_gemma2-9b-it}
\end{center}
\end{figure}

\begin{figure}[!ht]
\begin{center}
\ContinuedFloat
\captionsetup{list=off,format=cont}
\centering
    \setcounter{subfigure}{2}
    \subfigure[The $\text{3}^\text{rd}$ head]{\includegraphics[width=0.83\columnwidth]{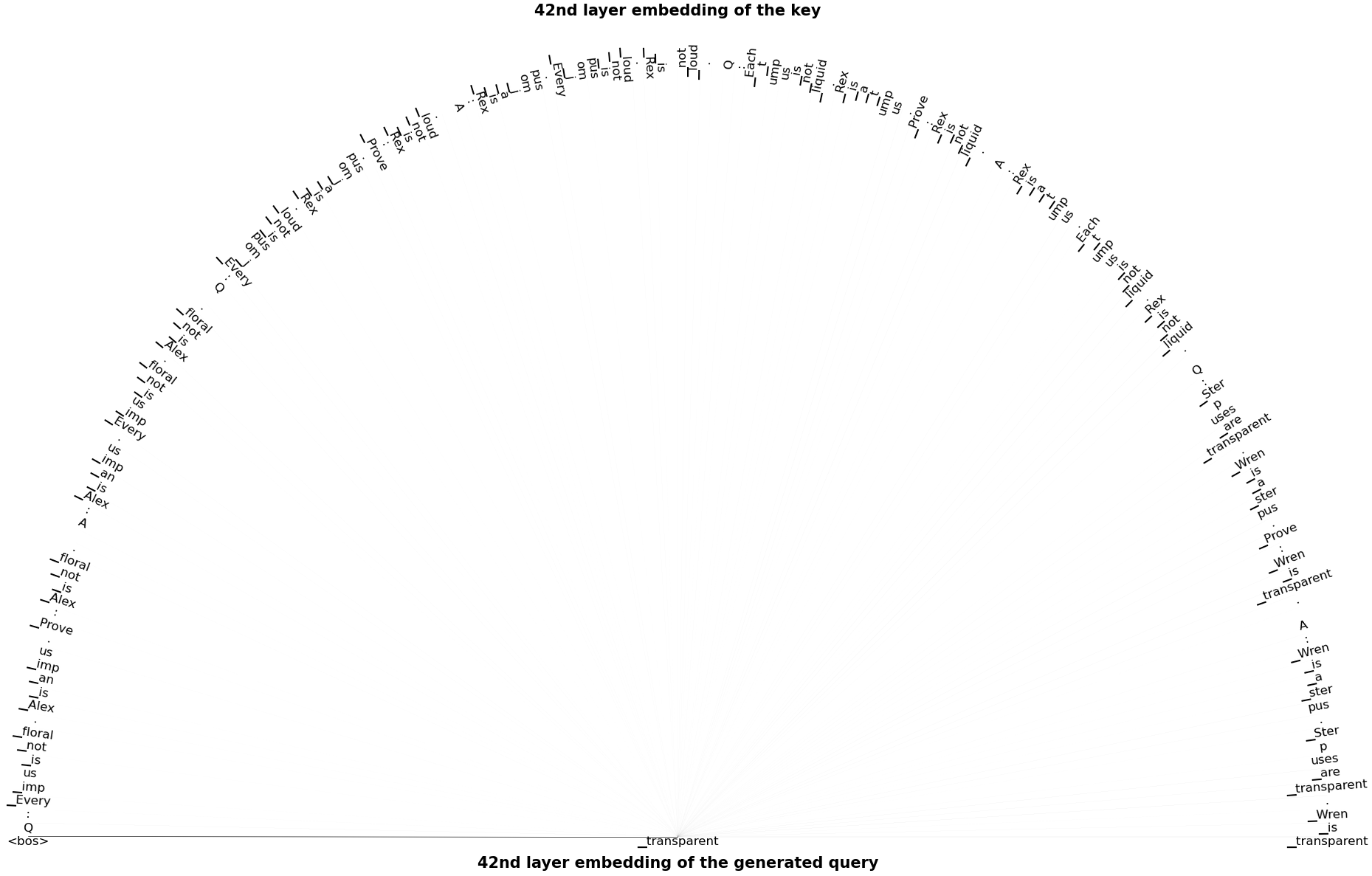}\label{fig:gemma2_H3}}
    \subfigure[The $\text{4}^\text{th}$ head]{\includegraphics[width=0.83\columnwidth]{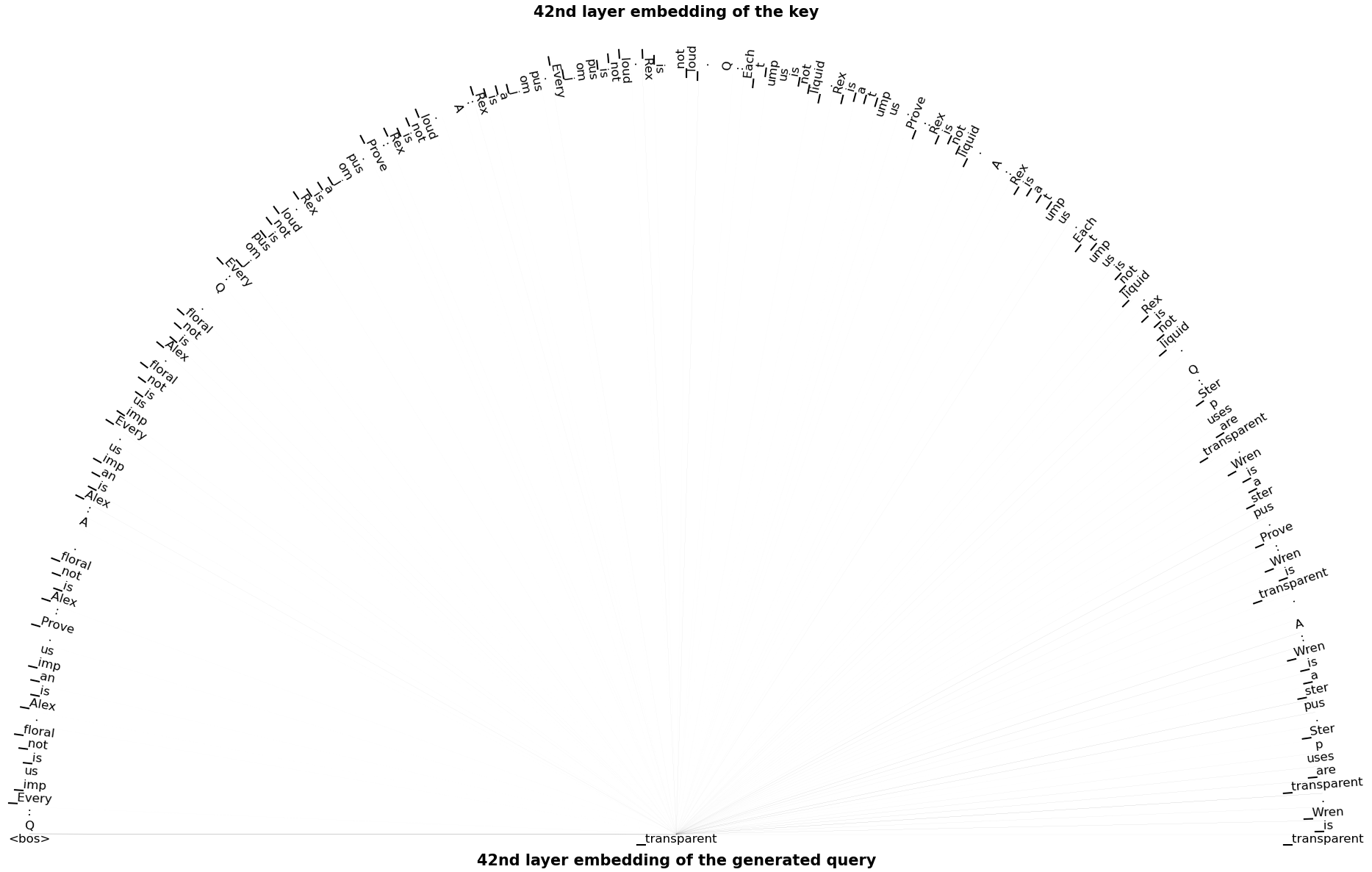}\label{fig:gemma2_H4}}
    \caption{Visualization of attention maps for each head in the final attention layer of the gemma2-9b-it model loaded with float16 precision. The model is prompted using the CoT prompt for the first proof of the implication elimination task from the PrOntoQA-OOD dataset \citep{saparov2023testing}, appended with the text: "Wren is a sterpus. Sterpuses are transparent. Wren is transparent". The thickness of each line connecting token embeddings is proportional to the corresponding attention score. The arc-shaped text represents the key embeddings, progressing from the bottom-left to bottom-right. The text at the bottom represents the query embedding of the last token.}
\label{Figure_ProntoQA_MP_attention_gemma2-9b-it}
\end{center}
\end{figure}

\begin{figure}[!ht]
\begin{center}
\ContinuedFloat
\captionsetup{list=off,format=cont}
\centering
    \setcounter{subfigure}{4}
    \subfigure[The $\text{5}^\text{th}$ head]{\includegraphics[width=0.83\columnwidth]{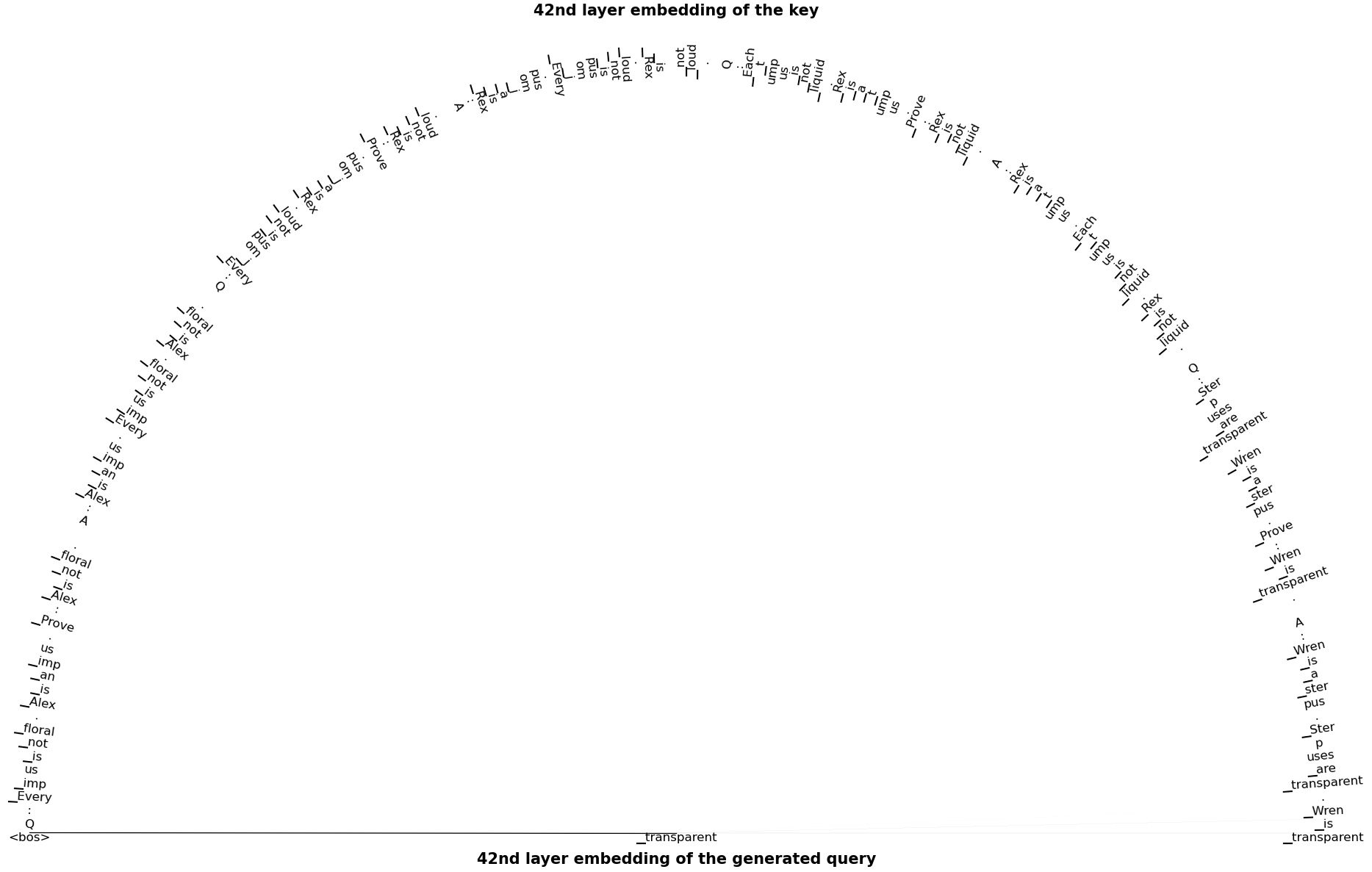}\label{fig:gemma2_H5}}
    \subfigure[The $\text{6}^\text{th}$ head]{\includegraphics[width=0.83\columnwidth]{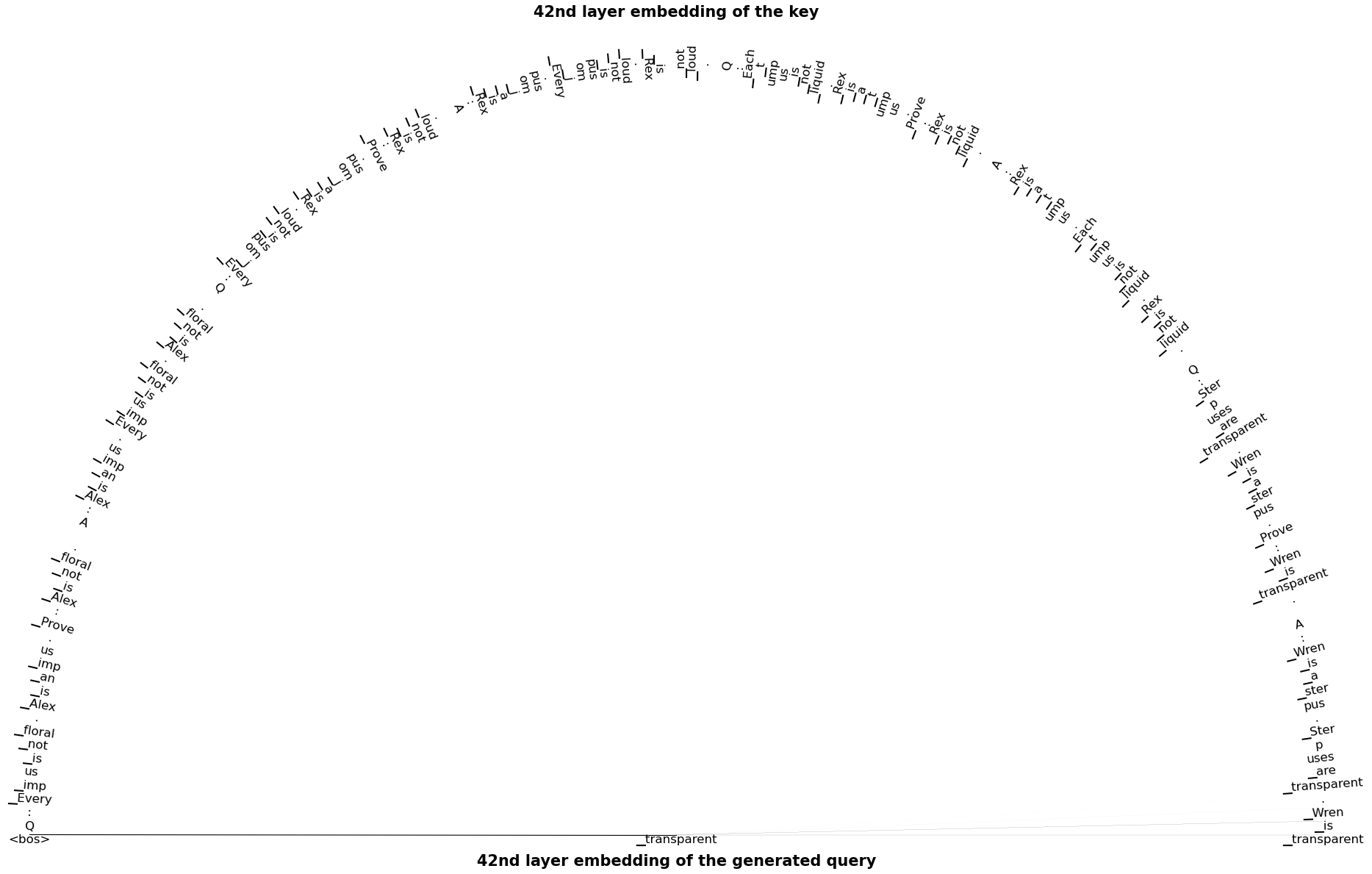}\label{fig:gemma2_H6}}
    \caption{Visualization of attention maps for each head in the final attention layer of the gemma2-9b-it model loaded with float16 precision. The model is prompted using the CoT prompt for the first proof of the implication elimination task from the PrOntoQA-OOD dataset \citep{saparov2023testing}, appended with the text: "Wren is a sterpus. Sterpuses are transparent. Wren is transparent". The thickness of each line connecting token embeddings is proportional to the corresponding attention score. The arc-shaped text represents the key embeddings, progressing from the bottom-left to bottom-right. The text at the bottom represents the query embedding of the last token.}
\label{Figure_ProntoQA_MP_attention_gemma2-9b-it}
\end{center}
\end{figure}

\begin{figure}[!ht]
\begin{center}
\ContinuedFloat
\captionsetup{list=off,format=cont}
\centering
    \setcounter{subfigure}{6}
    \subfigure[The $\text{7}^\text{th}$ head]{\includegraphics[width=0.83\columnwidth]{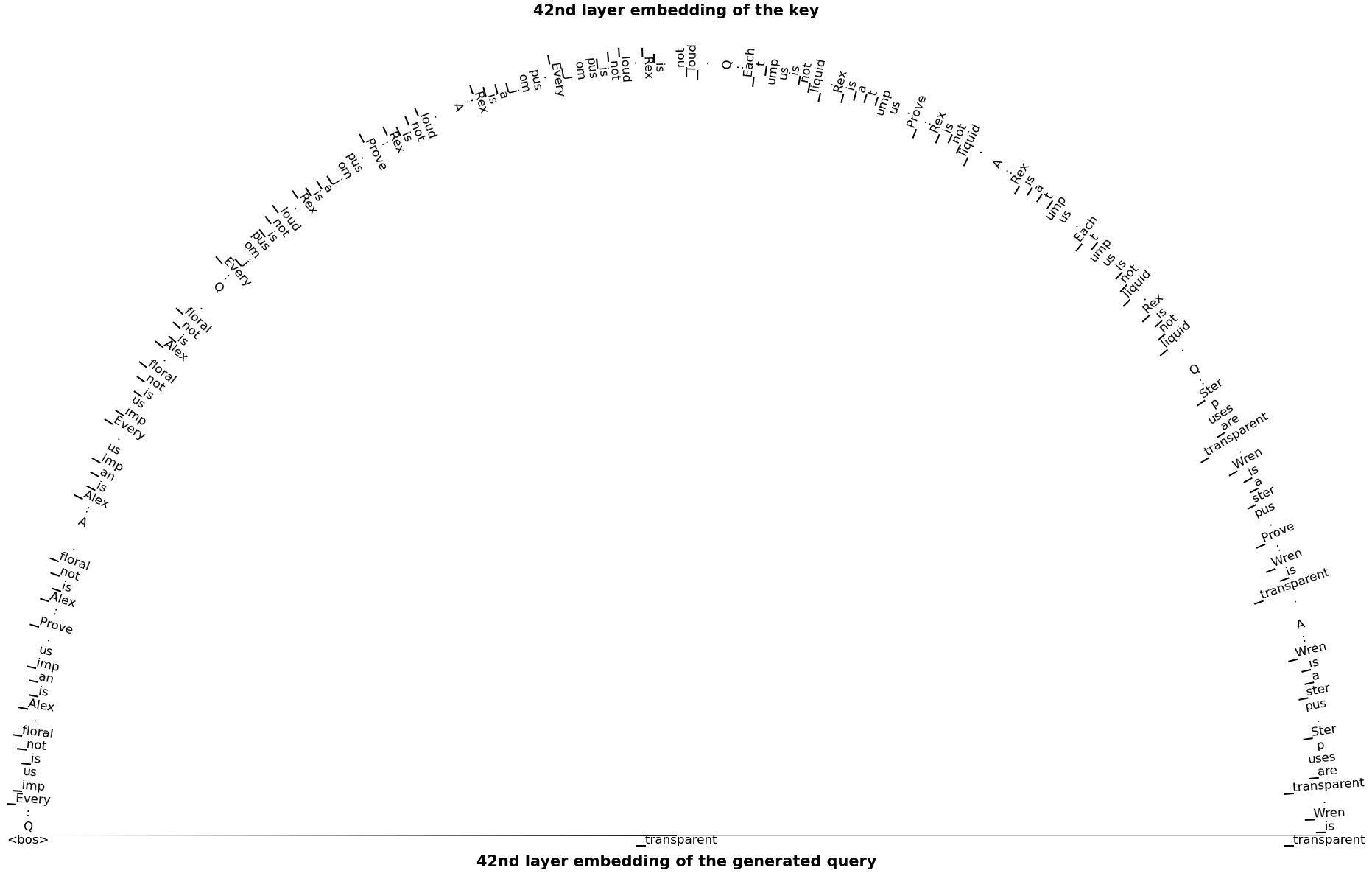}\label{fig:gemma2_H7}}
    \subfigure[The $\text{8}^\text{th}$ head]{\includegraphics[width=0.83\columnwidth]{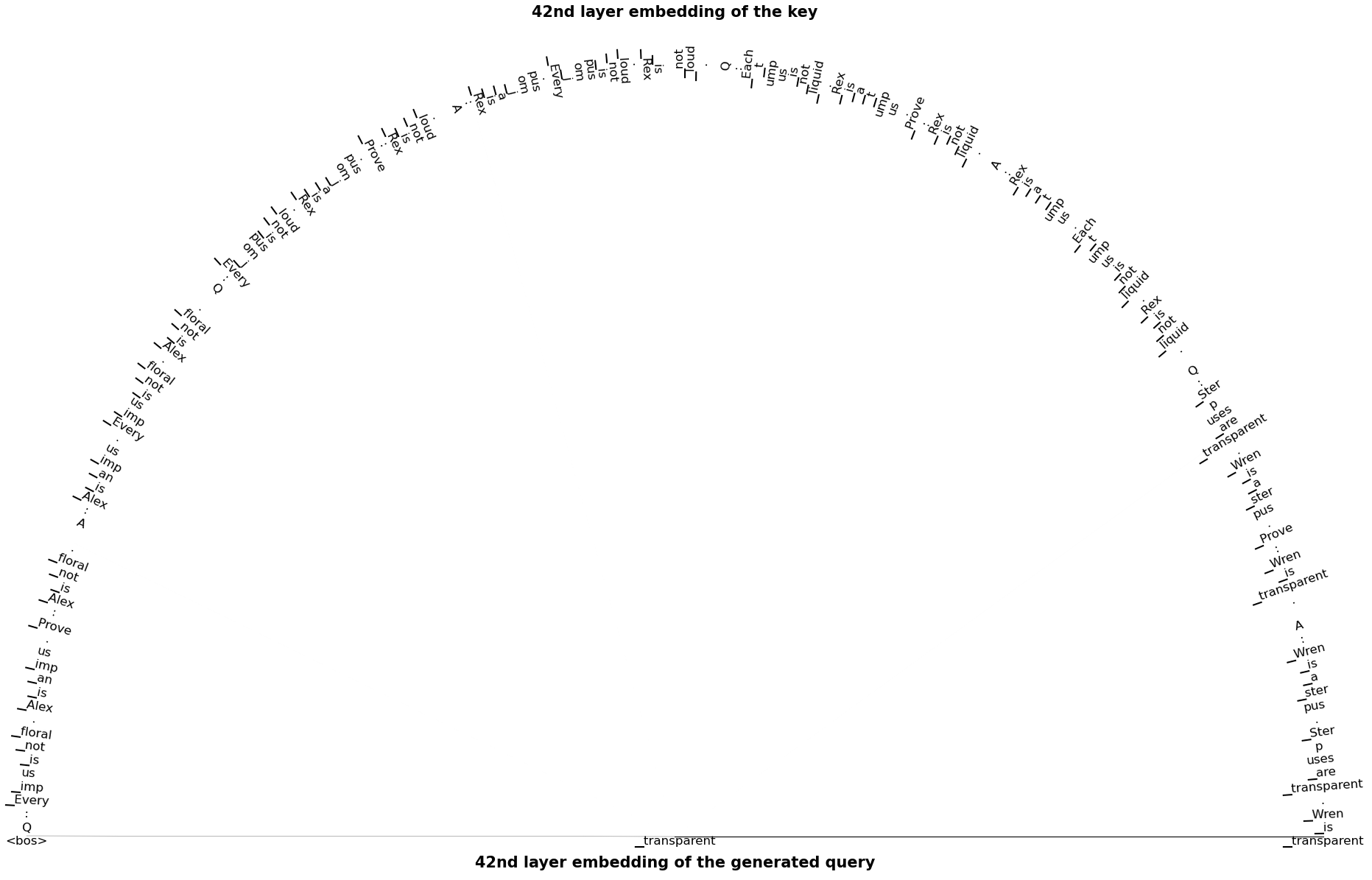}\label{fig:gemma2_H8}}
    \caption{Visualization of attention maps for each head in the final attention layer of the gemma2-9b-it model loaded with float16 precision. The model is prompted using the CoT prompt for the first proof of the implication elimination task from the PrOntoQA-OOD dataset \citep{saparov2023testing}, appended with the text: "Wren is a sterpus. Sterpuses are transparent. Wren is transparent". The thickness of each line connecting token embeddings is proportional to the corresponding attention score. The arc-shaped text represents the key embeddings, progressing from the bottom-left to bottom-right. The text at the bottom represents the query embedding of the last token.}
\label{Figure_ProntoQA_MP_attention_gemma2-9b-it}
\end{center}
\end{figure}

\begin{figure}[!ht]
\begin{center}
\ContinuedFloat
\captionsetup{list=off,format=cont}
\centering
    \setcounter{subfigure}{8}
    \subfigure[The $\text{9}^\text{th}$ head]{\includegraphics[width=0.83\columnwidth]{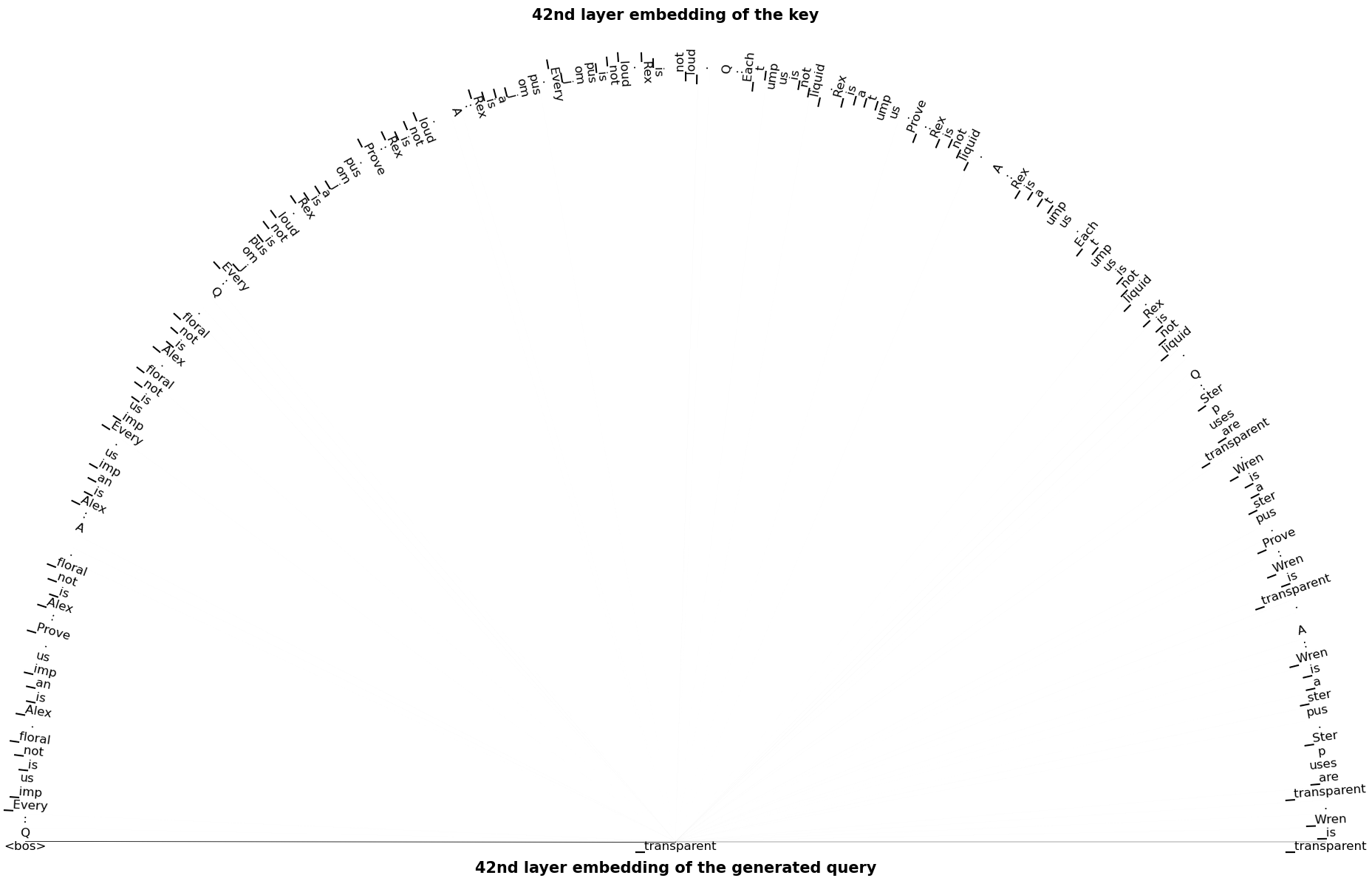}\label{fig:gemma2_H9}}
    \subfigure[The $\text{10}^\text{th}$ head]{\includegraphics[width=0.83\columnwidth]{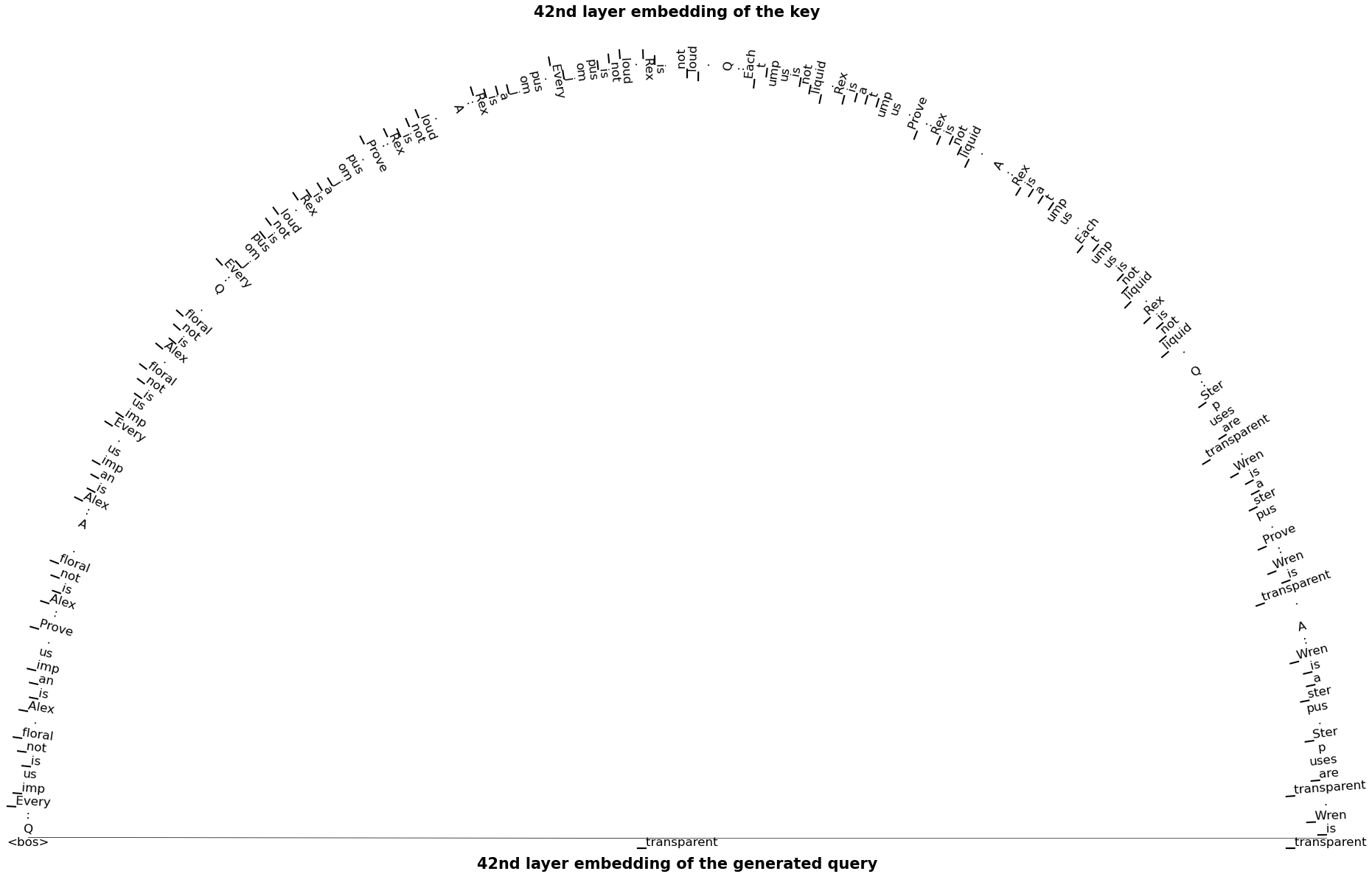}\label{fig:gemma2_H10}}
    \caption{Visualization of attention maps for each head in the final attention layer of the gemma2-9b-it model loaded with float16 precision. The model is prompted using the CoT prompt for the first proof of the implication elimination task from the PrOntoQA-OOD dataset \citep{saparov2023testing}, appended with the text: "Wren is a sterpus. Sterpuses are transparent. Wren is transparent". The thickness of each line connecting token embeddings is proportional to the corresponding attention score. The arc-shaped text represents the key embeddings, progressing from the bottom-left to bottom-right. The text at the bottom represents the query embedding of the last token.}
\label{Figure_ProntoQA_MP_attention_gemma2-9b-it}
\end{center}
\end{figure}

\begin{figure}[!ht]
\begin{center}
\ContinuedFloat
\captionsetup{list=off,format=cont}
\centering
    \setcounter{subfigure}{10}
    \subfigure[The $\text{11}^\text{th}$ head]{\includegraphics[width=0.83\columnwidth]{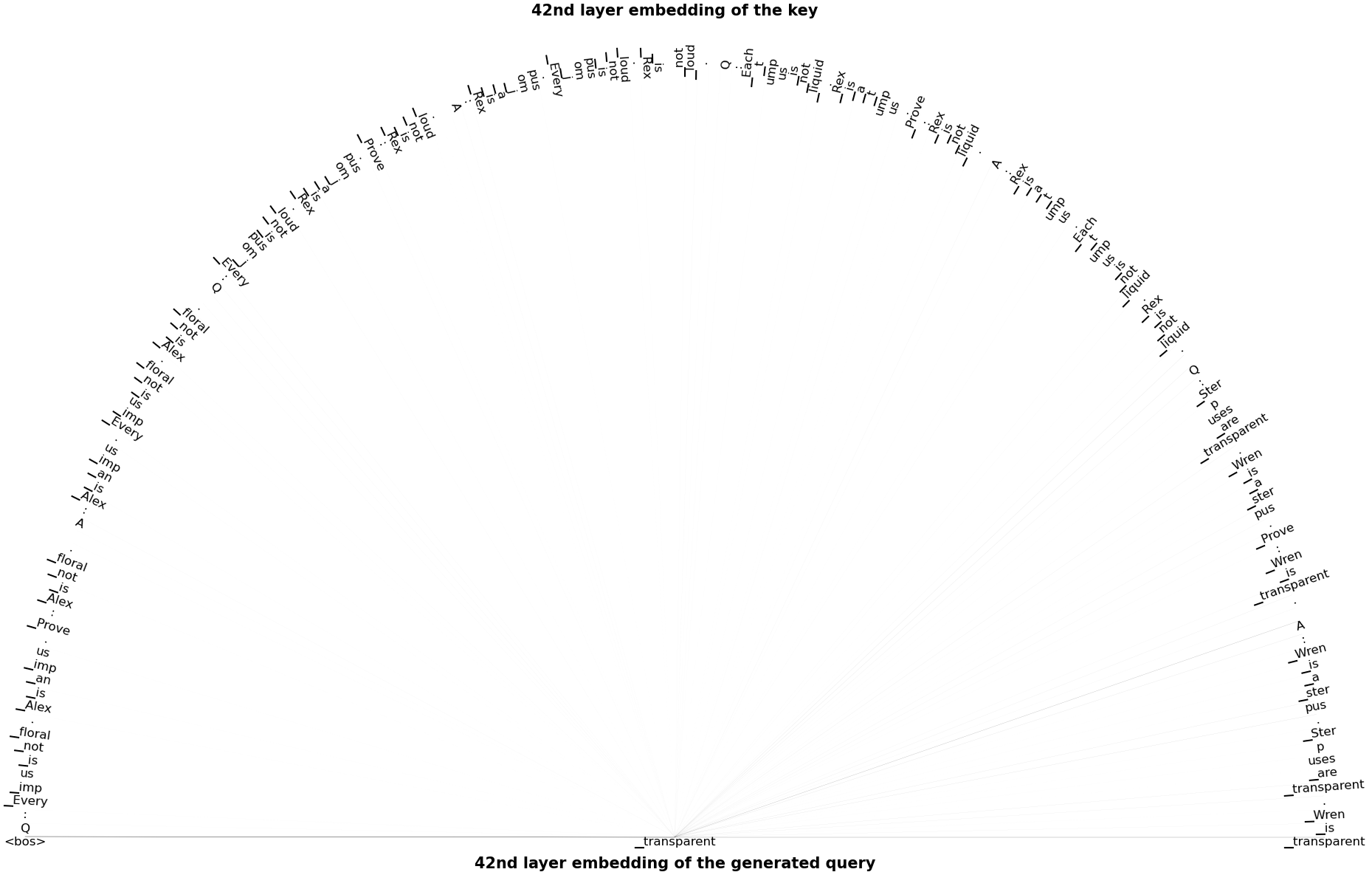}\label{fig:gemma2_H11}}
    \subfigure[The $\text{12}^\text{th}$ head]{\includegraphics[width=0.83\columnwidth]{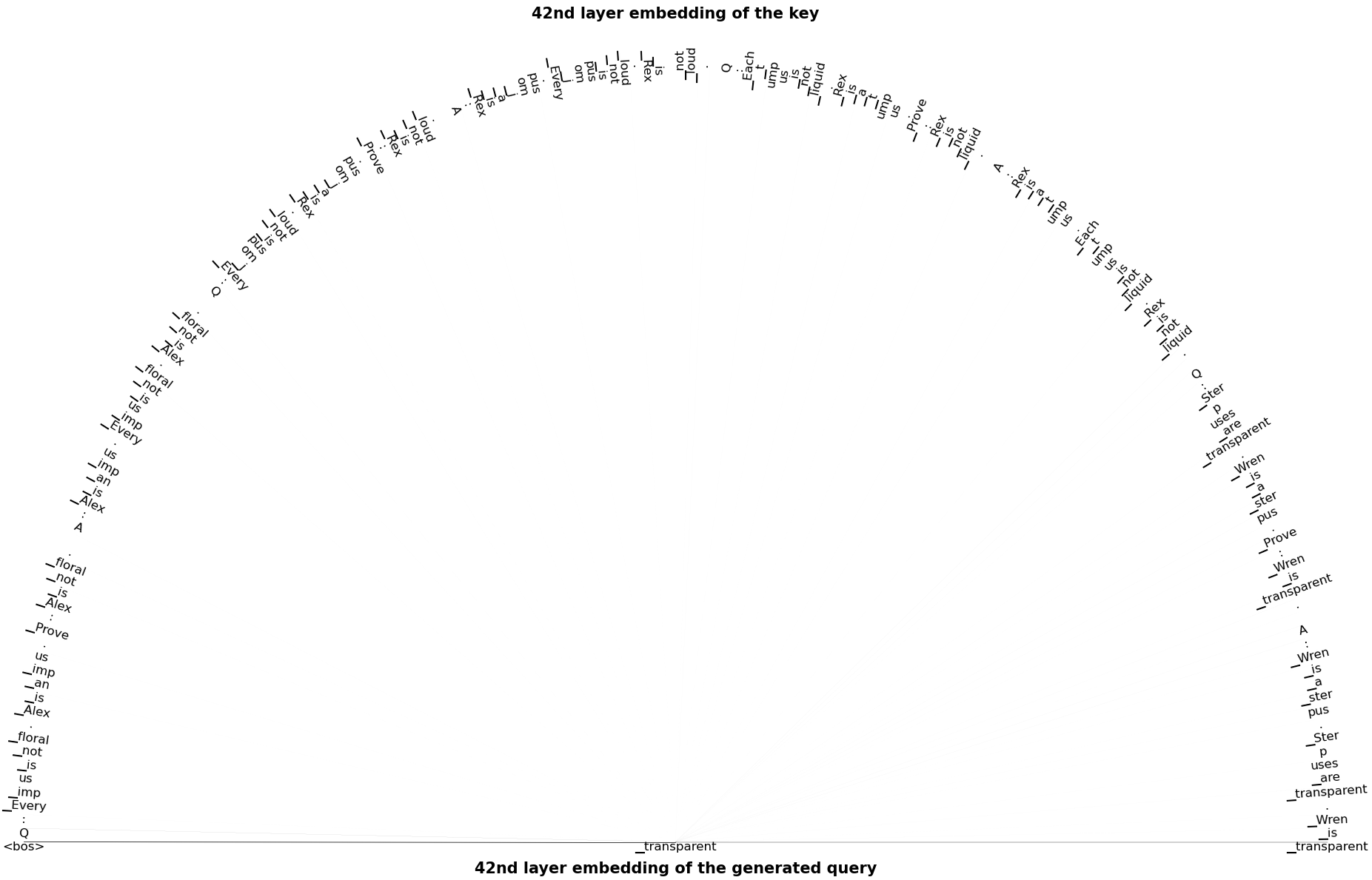}\label{fig:gemma2_H12}}
    \caption{Visualization of attention maps for each head in the final attention layer of the gemma2-9b-it model loaded with float16 precision. The model is prompted using the CoT prompt for the first proof of the implication elimination task from the PrOntoQA-OOD dataset \citep{saparov2023testing}, appended with the text: "Wren is a sterpus. Sterpuses are transparent. Wren is transparent". The thickness of each line connecting token embeddings is proportional to the corresponding attention score. The arc-shaped text represents the key embeddings, progressing from the bottom-left to bottom-right. The text at the bottom represents the query embedding of the last token.}
\label{Figure_ProntoQA_MP_attention_gemma2-9b-it}
\end{center}
\end{figure}

\begin{figure}[!ht]
\begin{center}
\ContinuedFloat
\captionsetup{list=off,format=cont}
\centering
    \setcounter{subfigure}{12}
    \subfigure[The $\text{13}^\text{th}$ head]{\includegraphics[width=0.83\columnwidth]{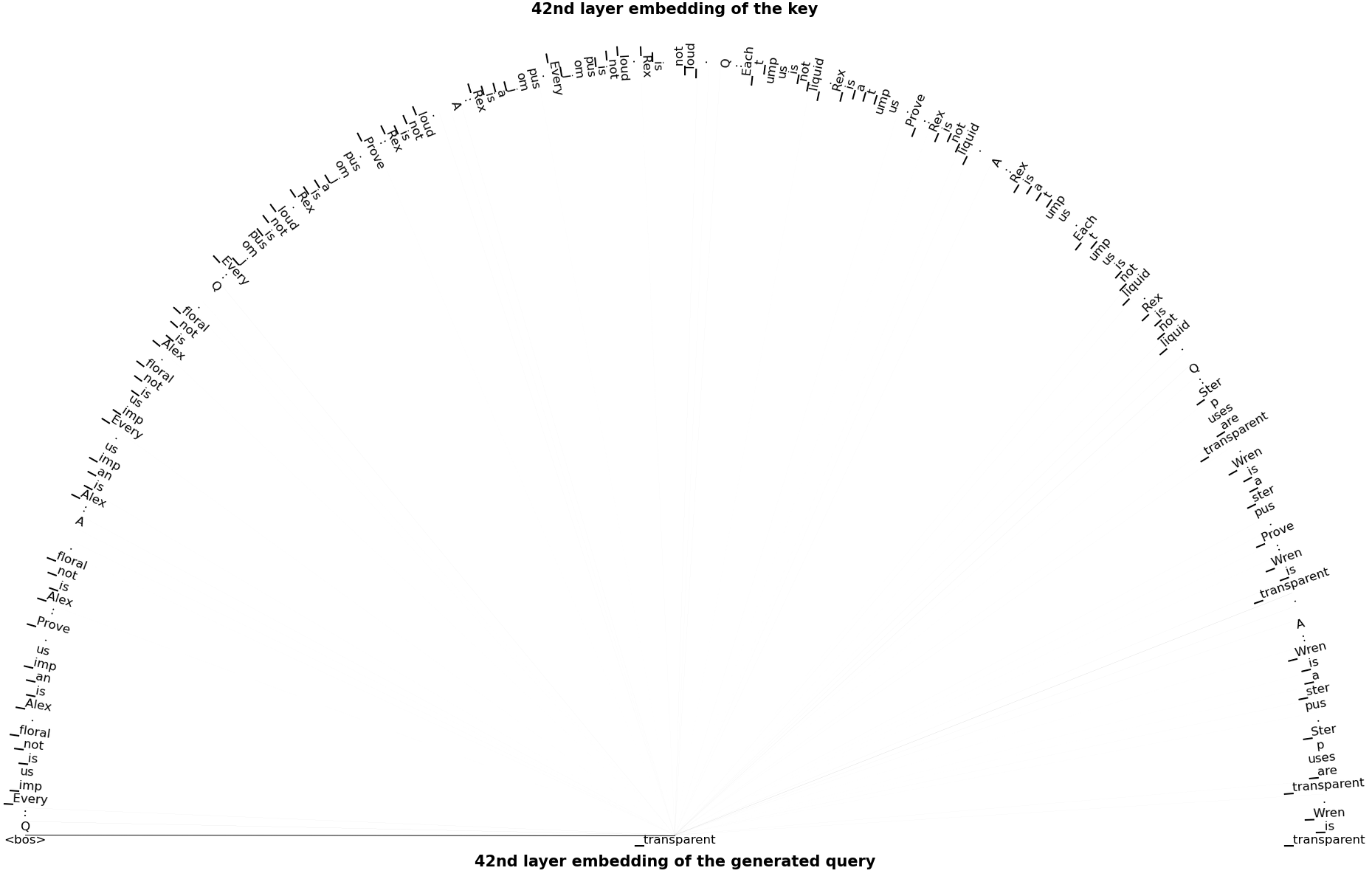}\label{fig:gemma2_H13}}
    \subfigure[The $\text{14}^\text{th}$ head]{\includegraphics[width=0.83\columnwidth]{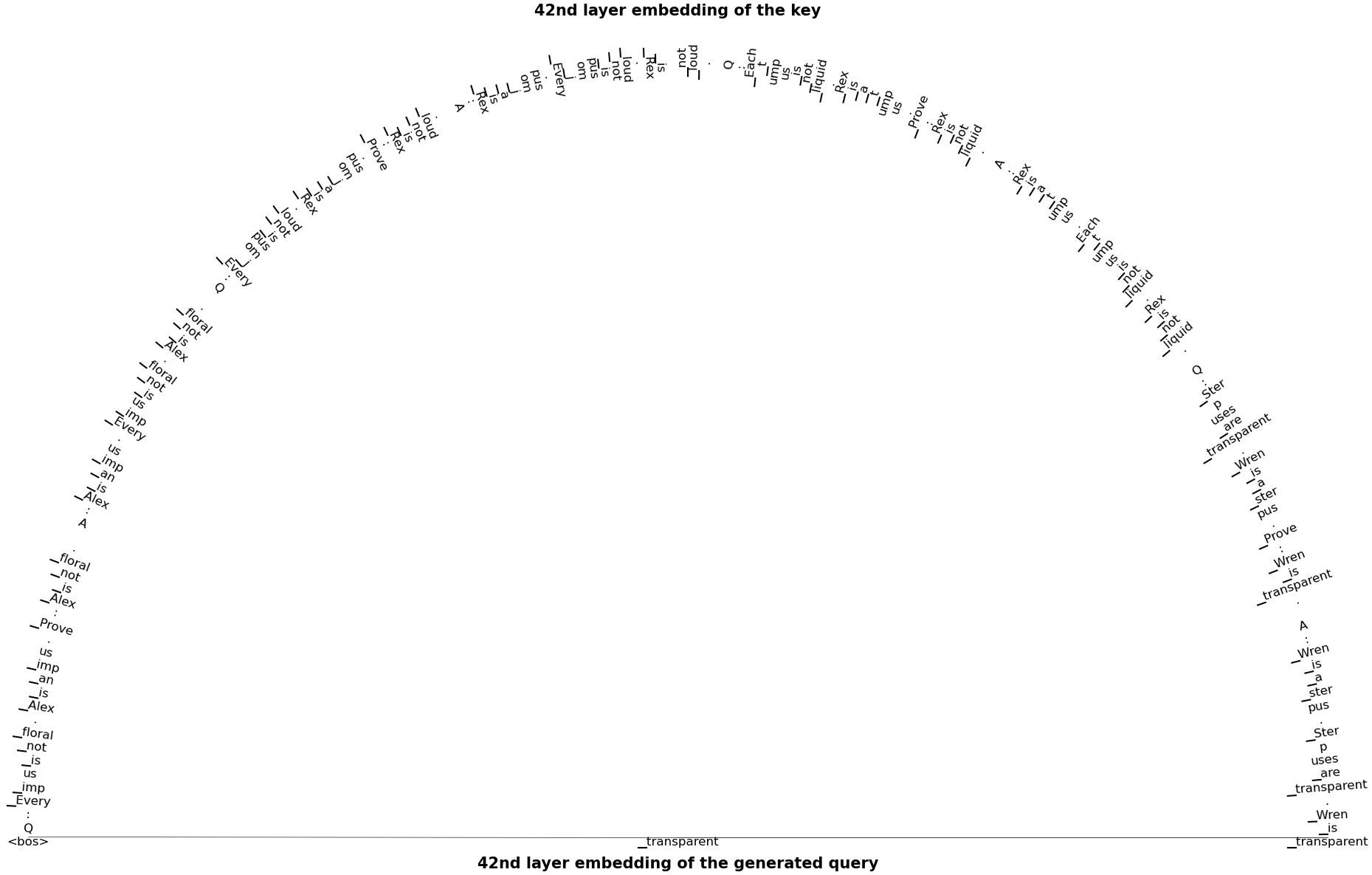}\label{fig:gemma2_H14}}
    \caption{Visualization of attention maps for each head in the final attention layer of the gemma2-9b-it model loaded with float16 precision. The model is prompted using the CoT prompt for the first proof of the implication elimination task from the PrOntoQA-OOD dataset \citep{saparov2023testing}, appended with the text: "Wren is a sterpus. Sterpuses are transparent. Wren is transparent". The thickness of each line connecting token embeddings is proportional to the corresponding attention score. The arc-shaped text represents the key embeddings, progressing from the bottom-left to bottom-right. The text at the bottom represents the query embedding of the last token.}
\label{Figure_ProntoQA_MP_attention_gemma2-9b-it}
\end{center}
\end{figure}

\begin{figure}[!ht]
\begin{center}
\ContinuedFloat
\captionsetup{list=off,format=cont}
\centering
    \setcounter{subfigure}{14}
    \subfigure[The $\text{15}^\text{th}$ head]{\includegraphics[width=0.83\columnwidth]{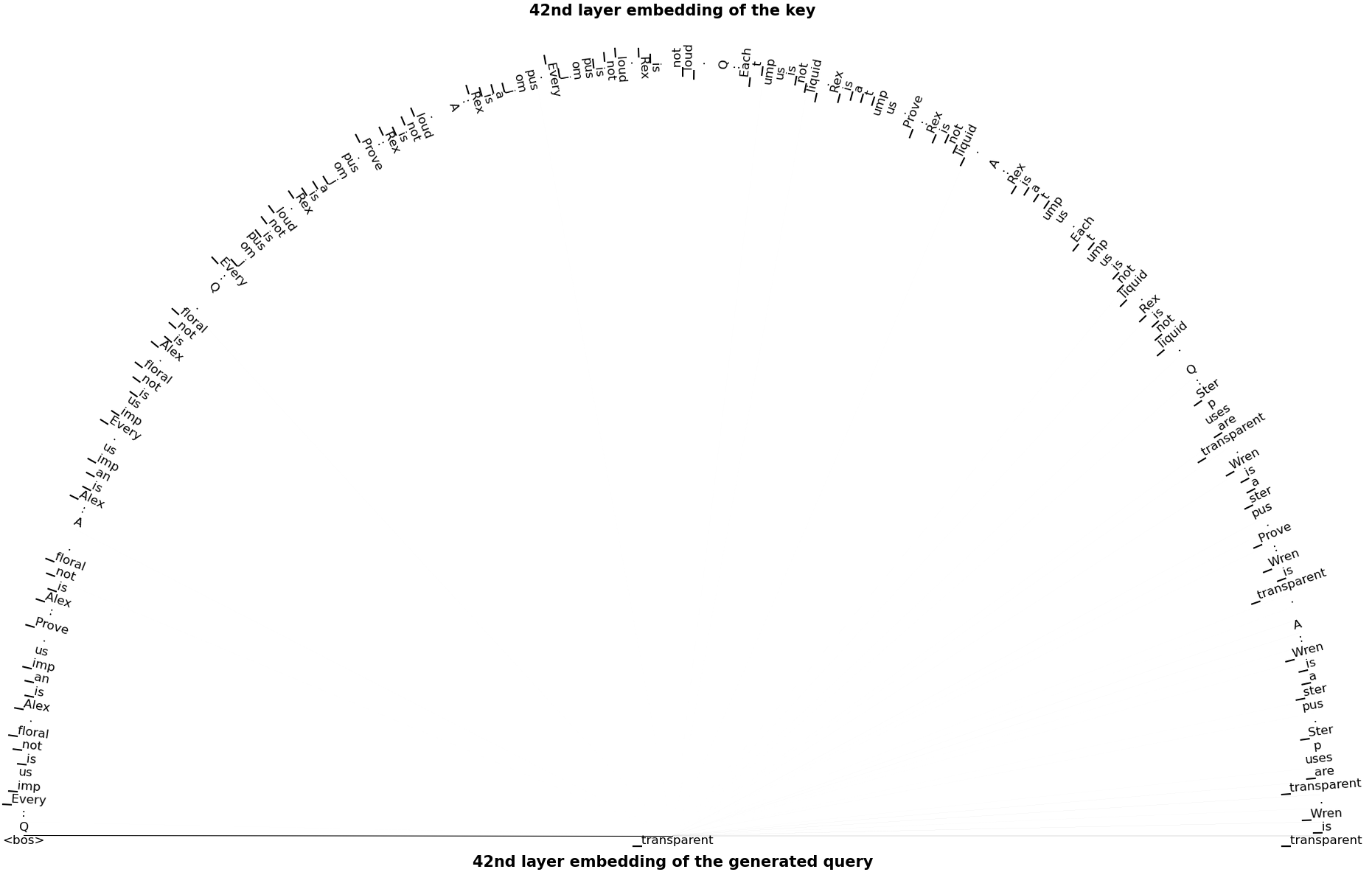}\label{fig:gemma2_H15}}
    \subfigure[The $\text{16}^\text{th}$ head]{\includegraphics[width=0.83\columnwidth]{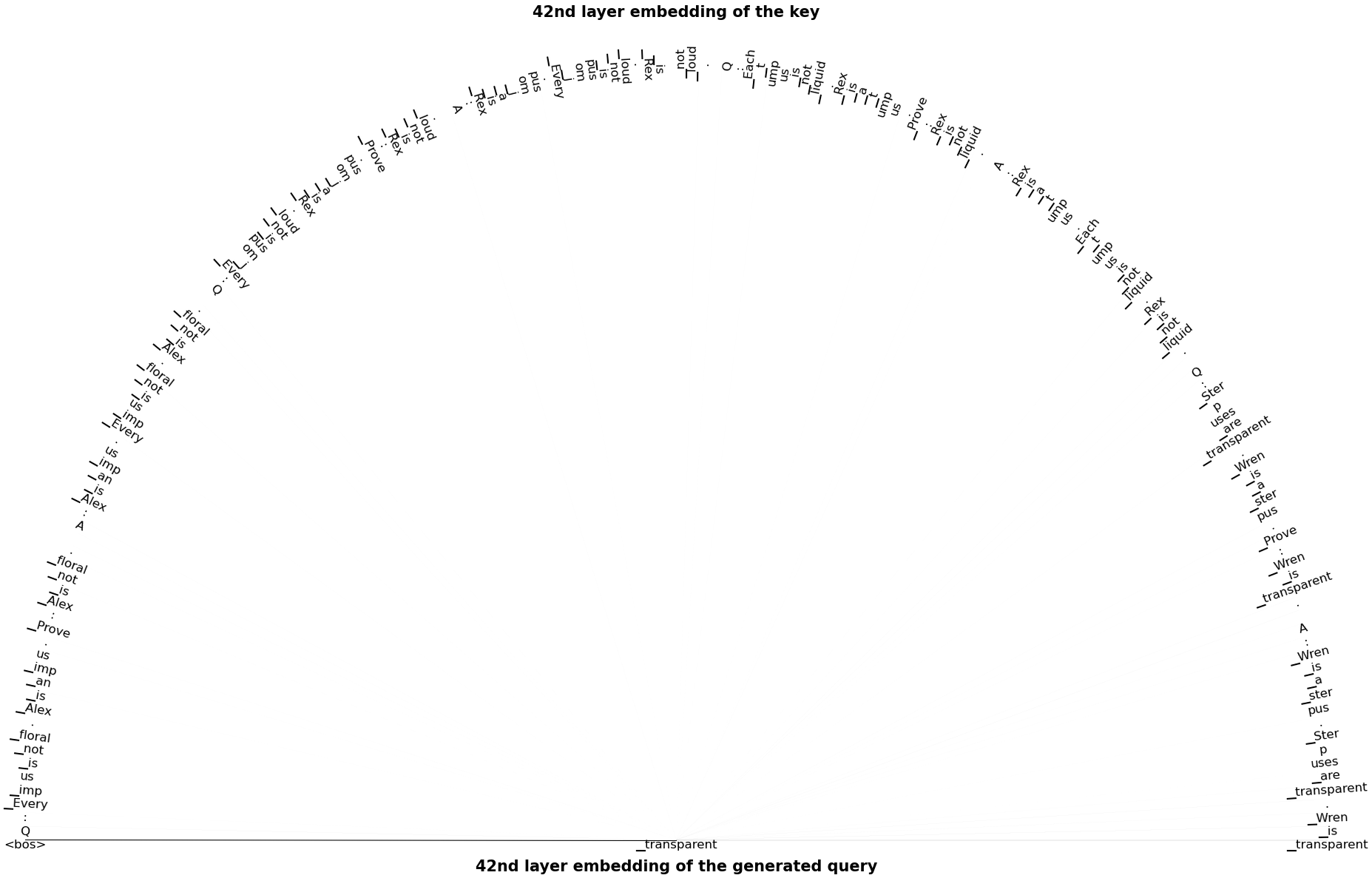}\label{fig:gemma2_H16}}
    \caption{Visualization of attention maps for each head in the final attention layer of the gemma2-9b-it model loaded with float16 precision. The model is prompted using the CoT prompt for the first proof of the implication elimination task from the PrOntoQA-OOD dataset \citep{saparov2023testing}, appended with the text: "Wren is a sterpus. Sterpuses are transparent. Wren is transparent". The thickness of each line connecting token embeddings is proportional to the corresponding attention score. The arc-shaped text represents the key embeddings, progressing from the bottom-left to bottom-right. The text at the bottom represents the query embedding of the last token.}
\label{Figure_ProntoQA_MP_attention_gemma2-9b-it}
\end{center}
\end{figure}


\begin{figure}[!ht]
\begin{center}
\centering
    \subfigure[The $\text{1}^\text{st}$ head]{\includegraphics[width=0.83\columnwidth]{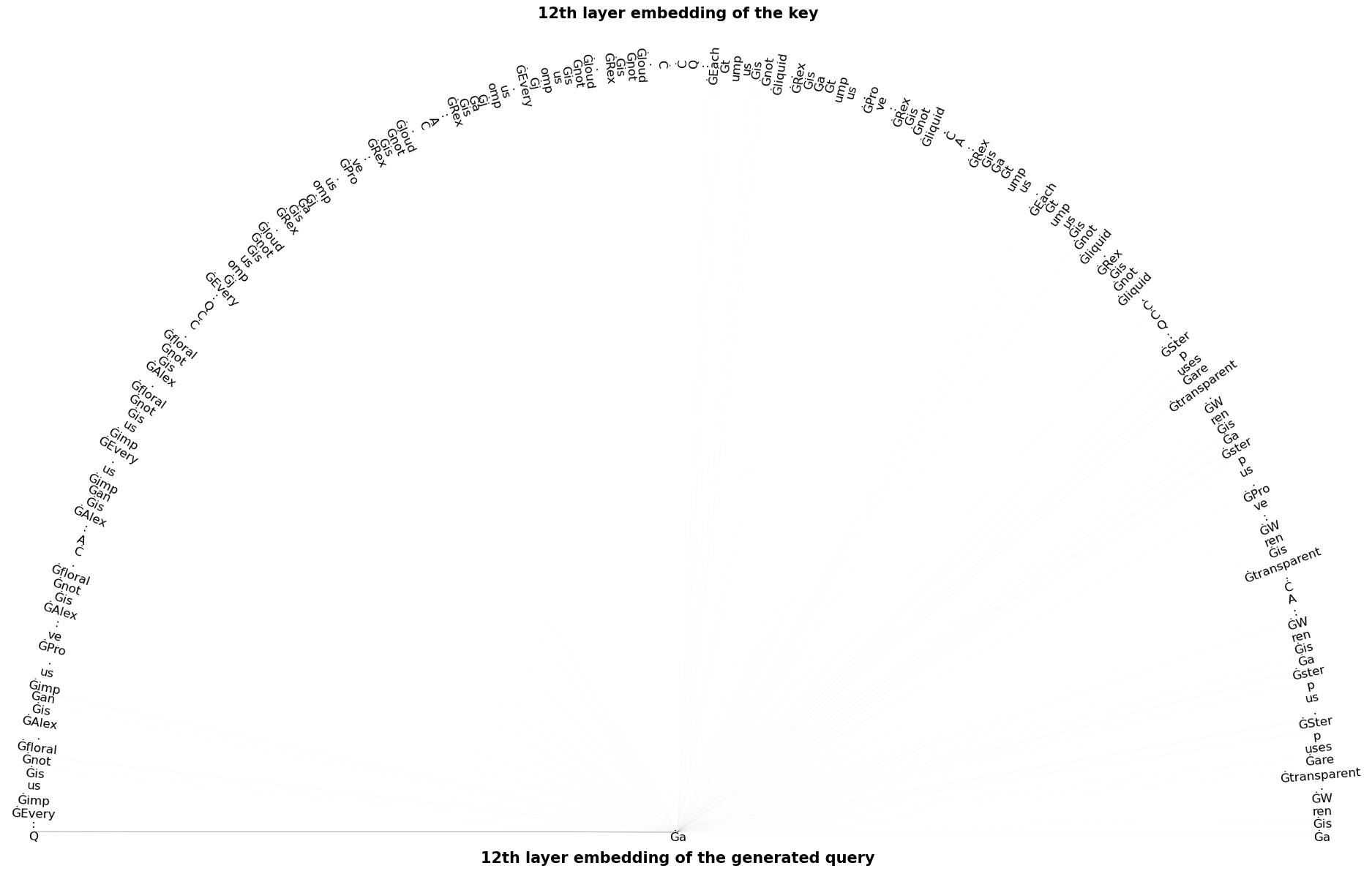}\label{fig:gpt2_H1}}
    \subfigure[The $\text{2}^\text{nd}$ head]{\includegraphics[width=0.83\columnwidth]{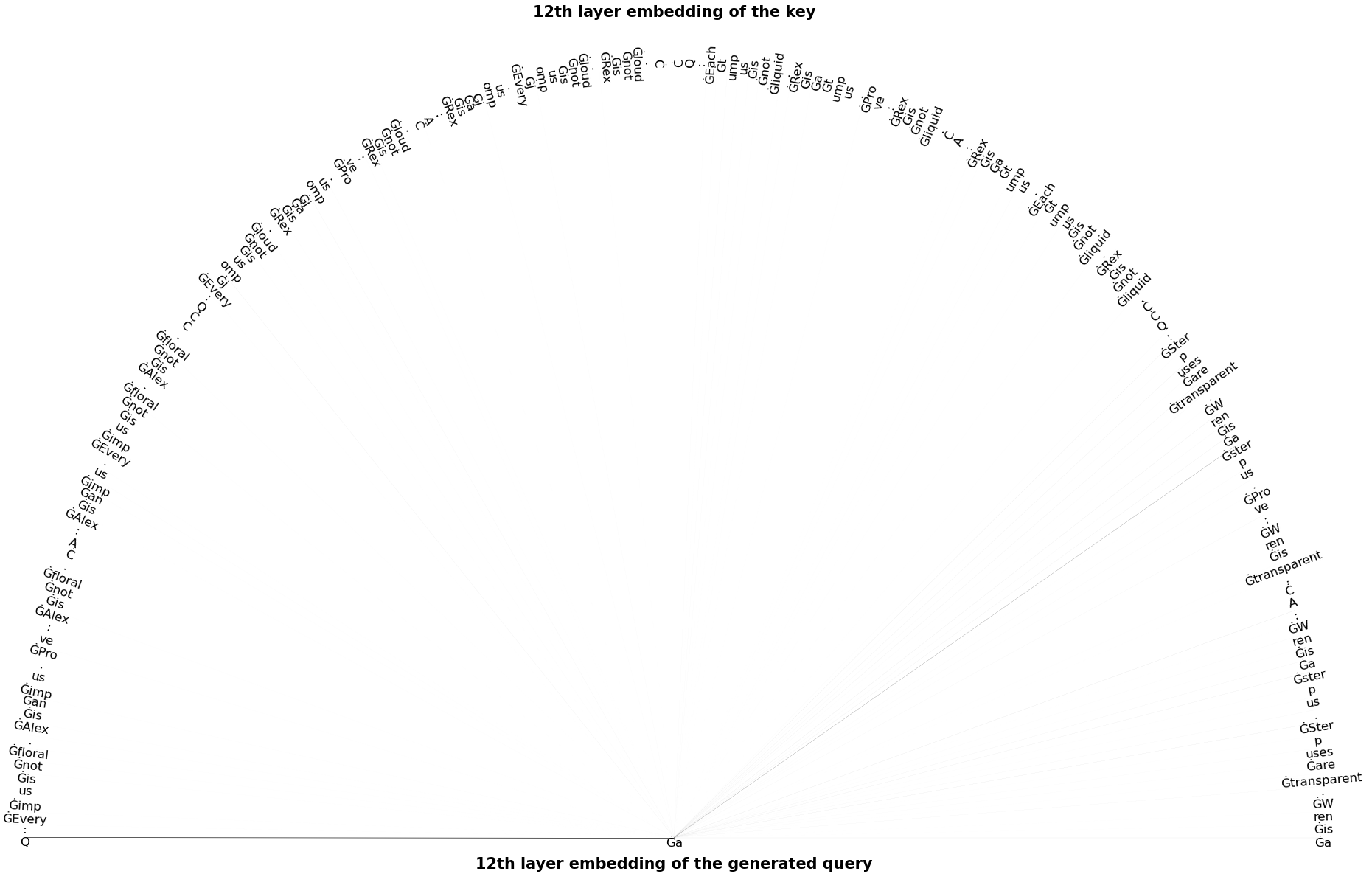}\label{fig:gpt2_H2}}
    \caption{Visualization of attention maps for each head in the final attention layer of the gpt2 model. The model is prompted using the CoT prompt for the first proof of the implication elimination task from the PrOntoQA-OOD dataset \citep{saparov2023testing}, appended with the text: "Wren is a sterpus. Sterpuses are transparent. Wren is a". The thickness of each line connecting token embeddings is proportional to the corresponding attention score. The arc-shaped text represents the key embeddings, progressing from the bottom-left to bottom-right. The text at the bottom represents the query embedding of the last token.}
\label{Figure_ProntoQA_MP_attention_gpt2}
\end{center}
\end{figure}

\begin{figure}[!ht]
\begin{center}
\ContinuedFloat
\captionsetup{list=off,format=cont}
\centering
    \setcounter{subfigure}{2}
    \subfigure[The $\text{3}^\text{rd}$ head]{\includegraphics[width=0.83\columnwidth]{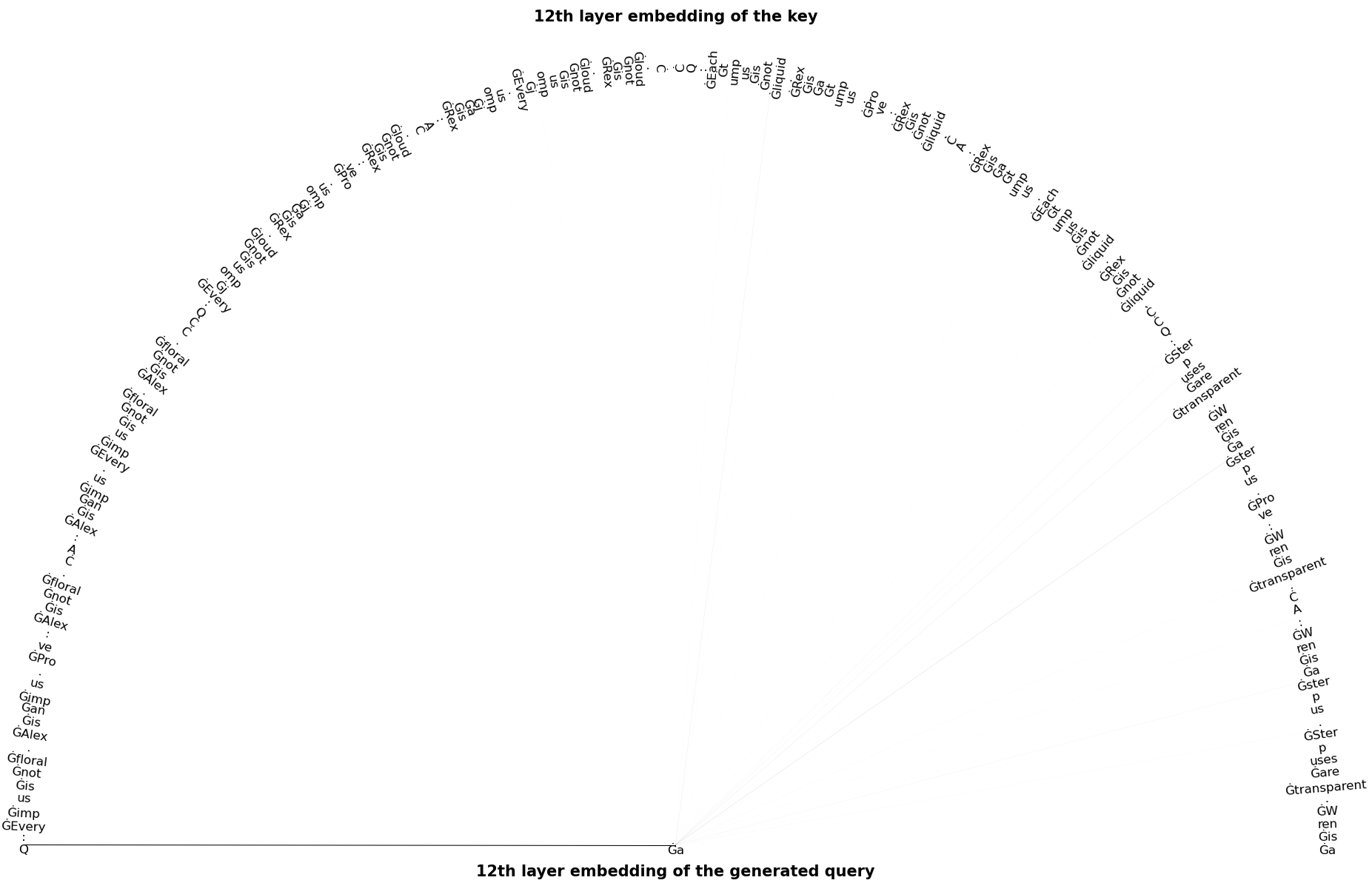}\label{fig:gpt2_H3}}
    \subfigure[The $\text{4}^\text{th}$ head]{\includegraphics[width=0.83\columnwidth]{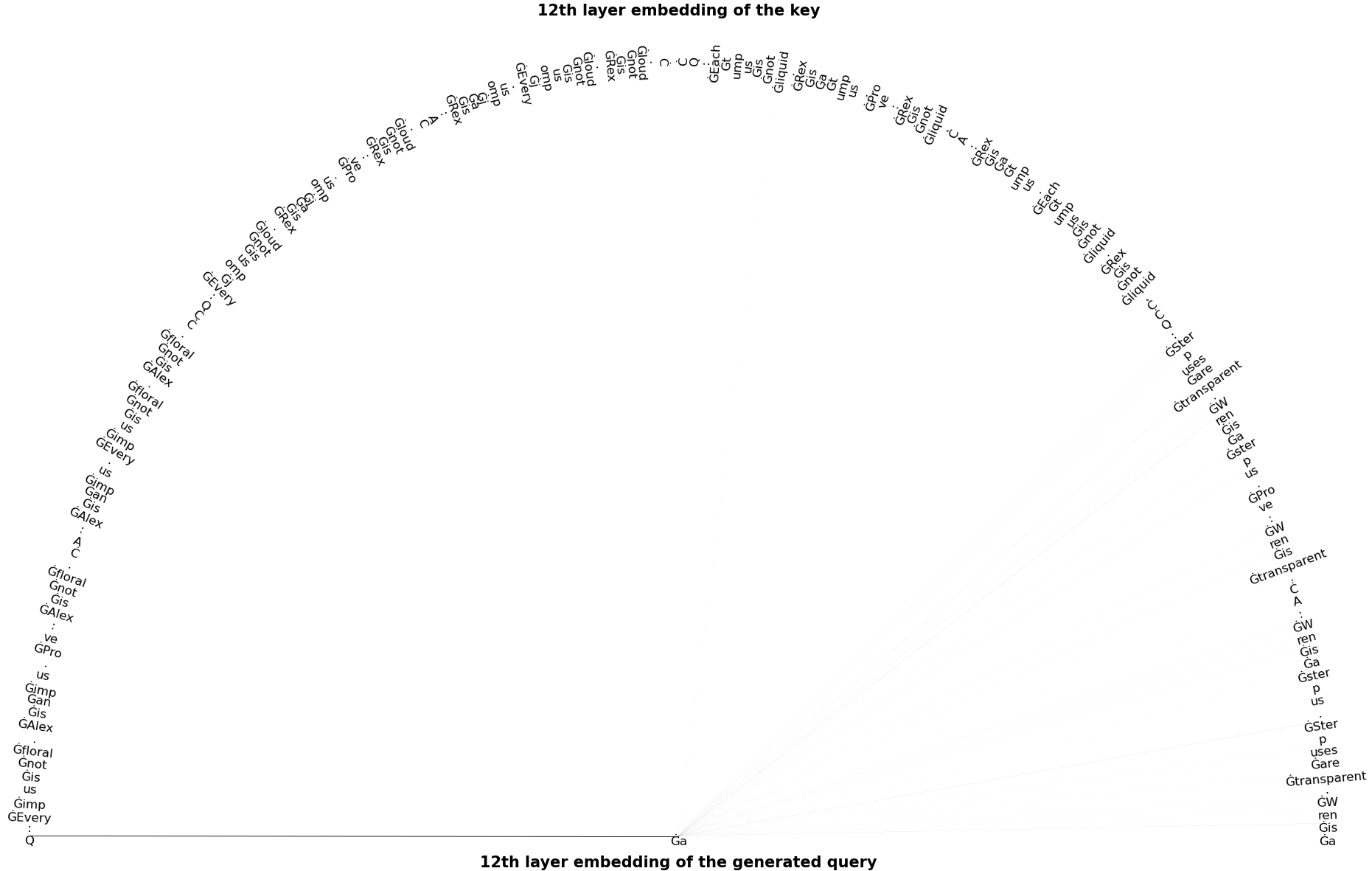}\label{fig:gpt2_H4}}
    \caption{Visualization of attention maps for each head in the final attention layer of the gemma2-9b-it model loaded with float16 precision. The model is prompted using the CoT prompt for the first proof of the implication elimination task from the PrOntoQA-OOD dataset \citep{saparov2023testing}, appended with the text: "Wren is a sterpus. Sterpuses are transparent. Wren is a". The thickness of each line connecting token embeddings is proportional to the corresponding attention score. The arc-shaped text represents the key embeddings, progressing from the bottom-left to bottom-right. The text at the bottom represents the query embedding of the last token.}
\label{Figure_ProntoQA_MP_attention_gpt2}
\end{center}
\end{figure}

\begin{figure}[!ht]
\begin{center}
\ContinuedFloat
\captionsetup{list=off,format=cont}
\centering
    \setcounter{subfigure}{4}
    \subfigure[The $\text{5}^\text{th}$ head]{\includegraphics[width=0.83\columnwidth]{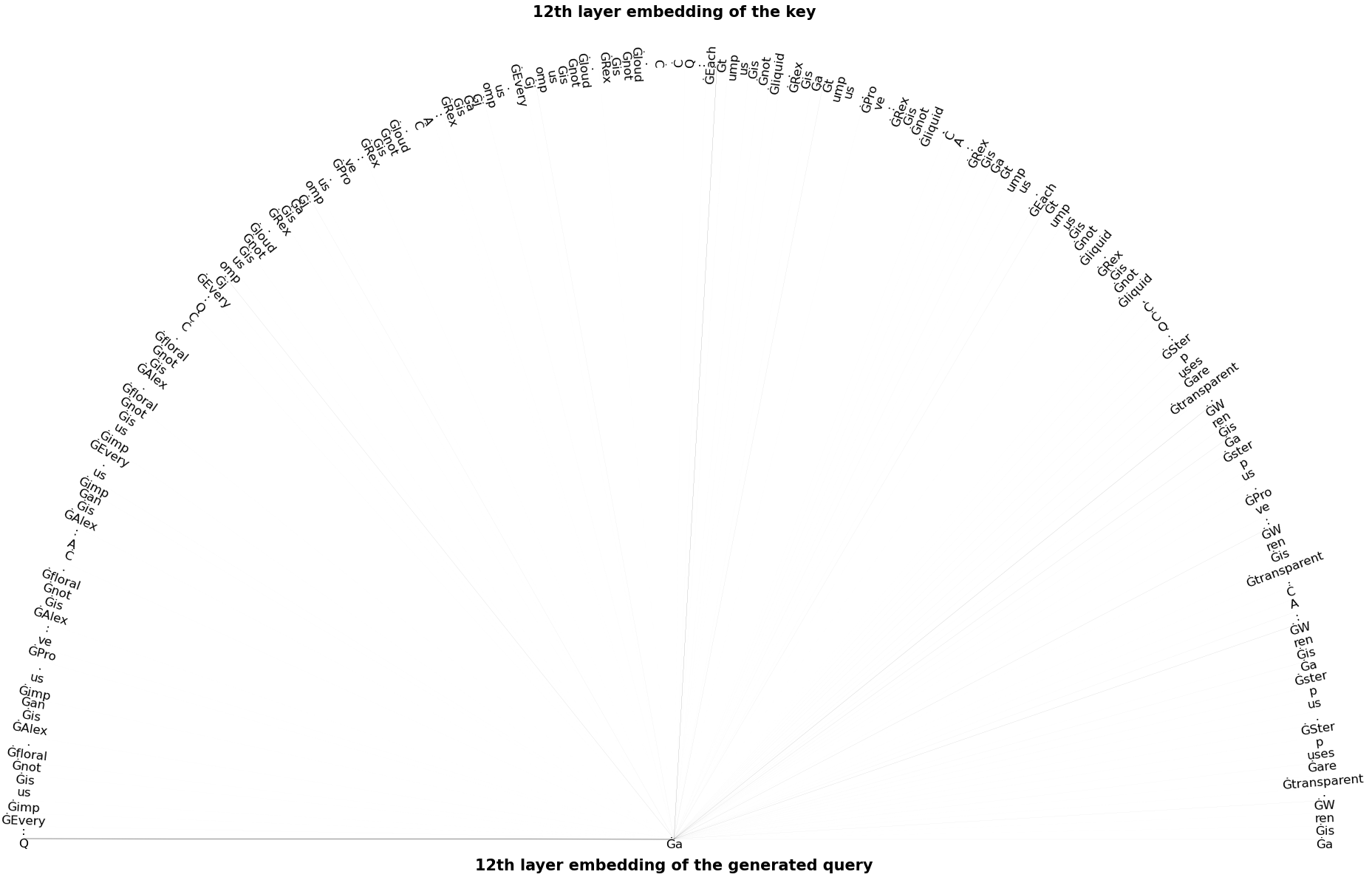}\label{fig:gpt2_H5}}
    \subfigure[The $\text{6}^\text{th}$ head]{\includegraphics[width=0.83\columnwidth]{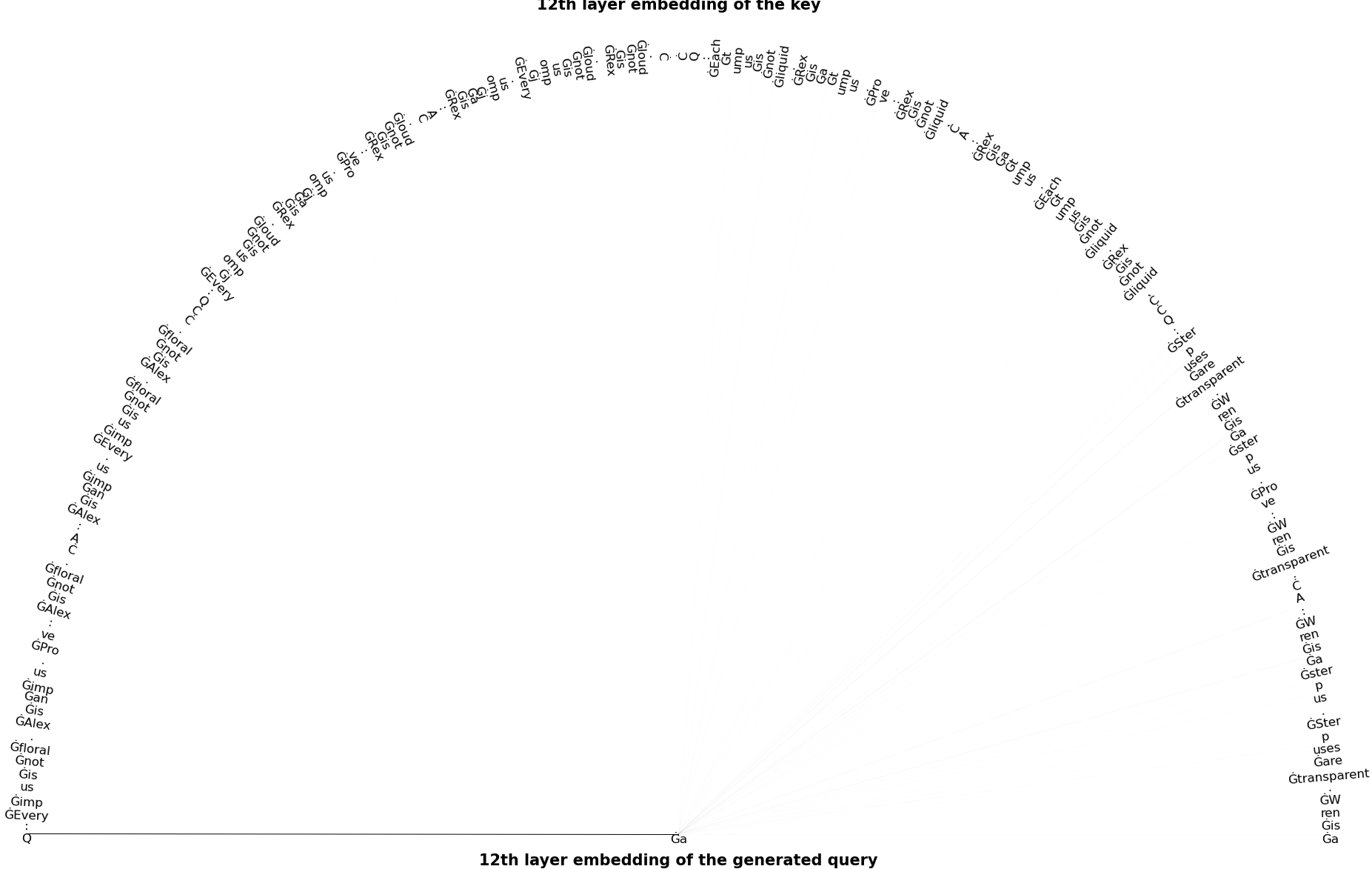}\label{fig:gpt2_H6}}
    \caption{Visualization of attention maps for each head in the final attention layer of the gemma2-9b-it model loaded with float16 precision. The model is prompted using the CoT prompt for the first proof of the implication elimination task from the PrOntoQA-OOD dataset \citep{saparov2023testing}, appended with the text: "Wren is a sterpus. Sterpuses are transparent. Wren is a". The thickness of each line connecting token embeddings is proportional to the corresponding attention score. The arc-shaped text represents the key embeddings, progressing from the bottom-left to bottom-right. The text at the bottom represents the query embedding of the last token.}
\label{Figure_ProntoQA_MP_attention_gpt2}
\end{center}
\end{figure}

\begin{figure}[!ht]
\begin{center}
\ContinuedFloat
\captionsetup{list=off,format=cont}
\centering
    \setcounter{subfigure}{6}
    \subfigure[The $\text{7}^\text{th}$ head]{\includegraphics[width=0.83\columnwidth]{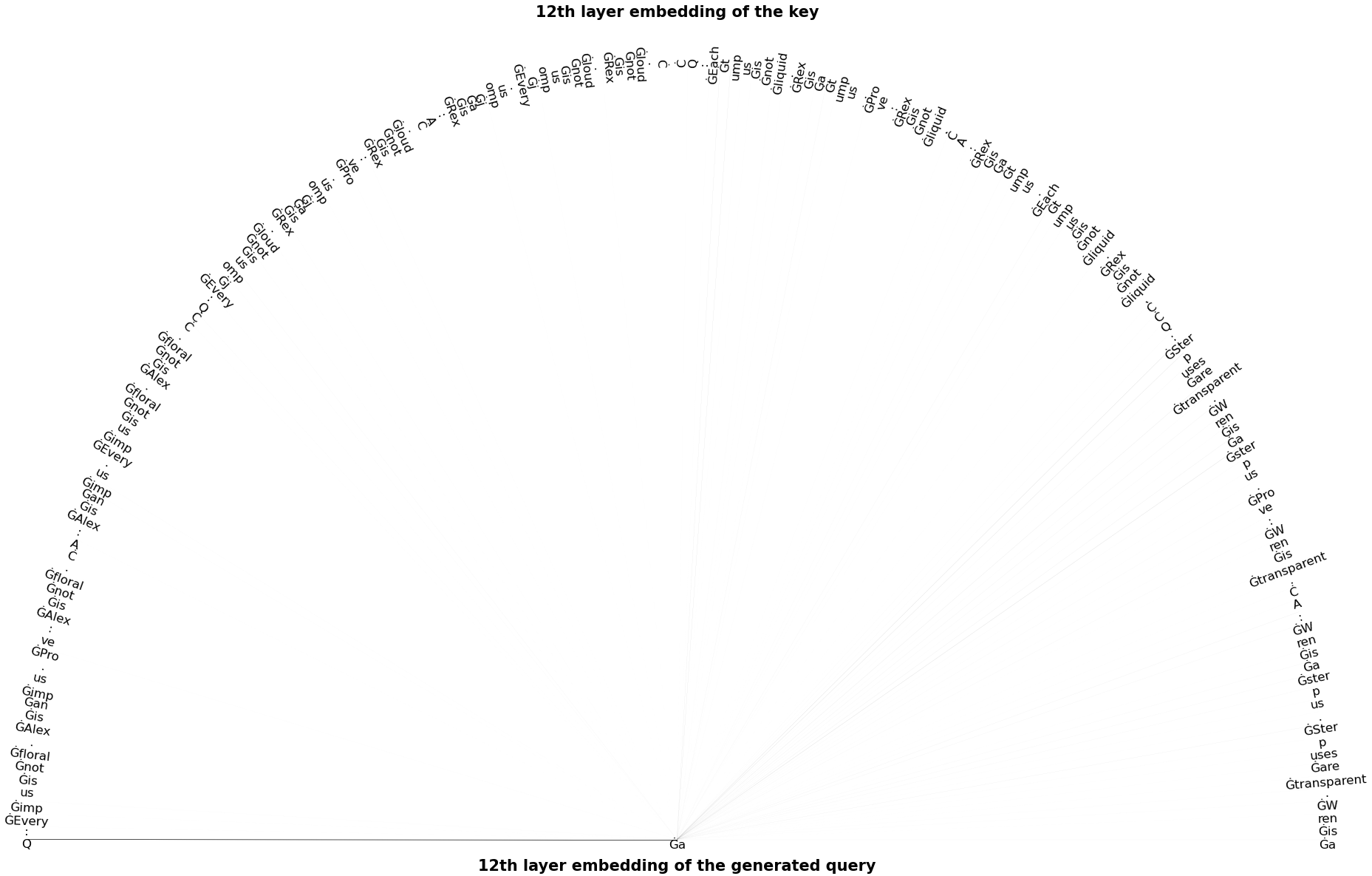}\label{fig:gpt2_H7}}
    \subfigure[The $\text{8}^\text{th}$ head]{\includegraphics[width=0.83\columnwidth]{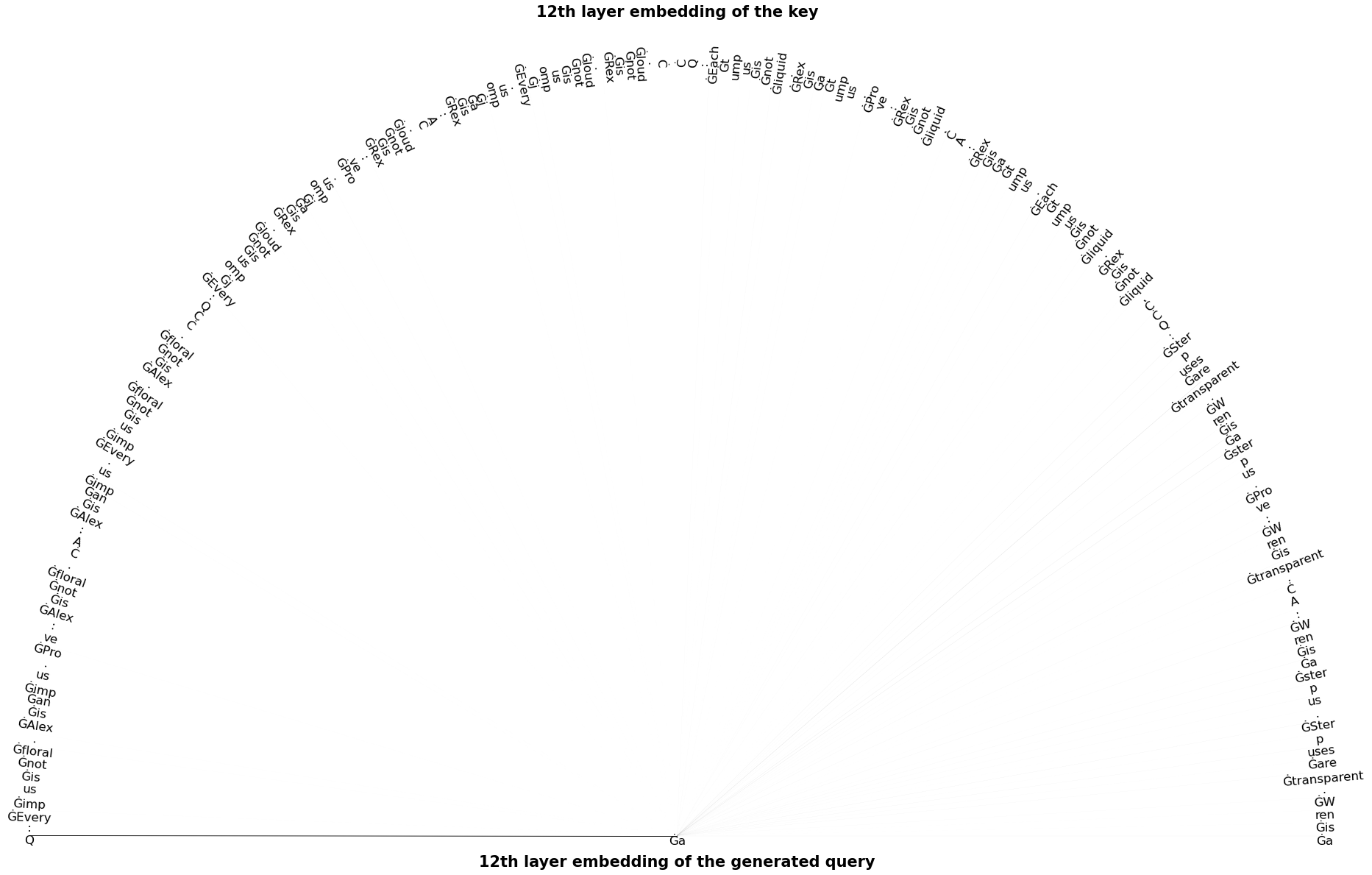}\label{fig:gpt2_H8}}
    \caption{Visualization of attention maps for each head in the final attention layer of the gemma2-9b-it model loaded with float16 precision. The model is prompted using the CoT prompt for the first proof of the implication elimination task from the PrOntoQA-OOD dataset \citep{saparov2023testing}, appended with the text: "Wren is a sterpus. Sterpuses are transparent. Wren is a". The thickness of each line connecting token embeddings is proportional to the corresponding attention score. The arc-shaped text represents the key embeddings, progressing from the bottom-left to bottom-right. The text at the bottom represents the query embedding of the last token.}
\label{Figure_ProntoQA_MP_attention_gpt2}
\end{center}
\end{figure}

\begin{figure}[!ht]
\begin{center}
\ContinuedFloat
\captionsetup{list=off,format=cont}
\centering
    \setcounter{subfigure}{8}
    \subfigure[The $\text{9}^\text{th}$ head]{\includegraphics[width=0.83\columnwidth]{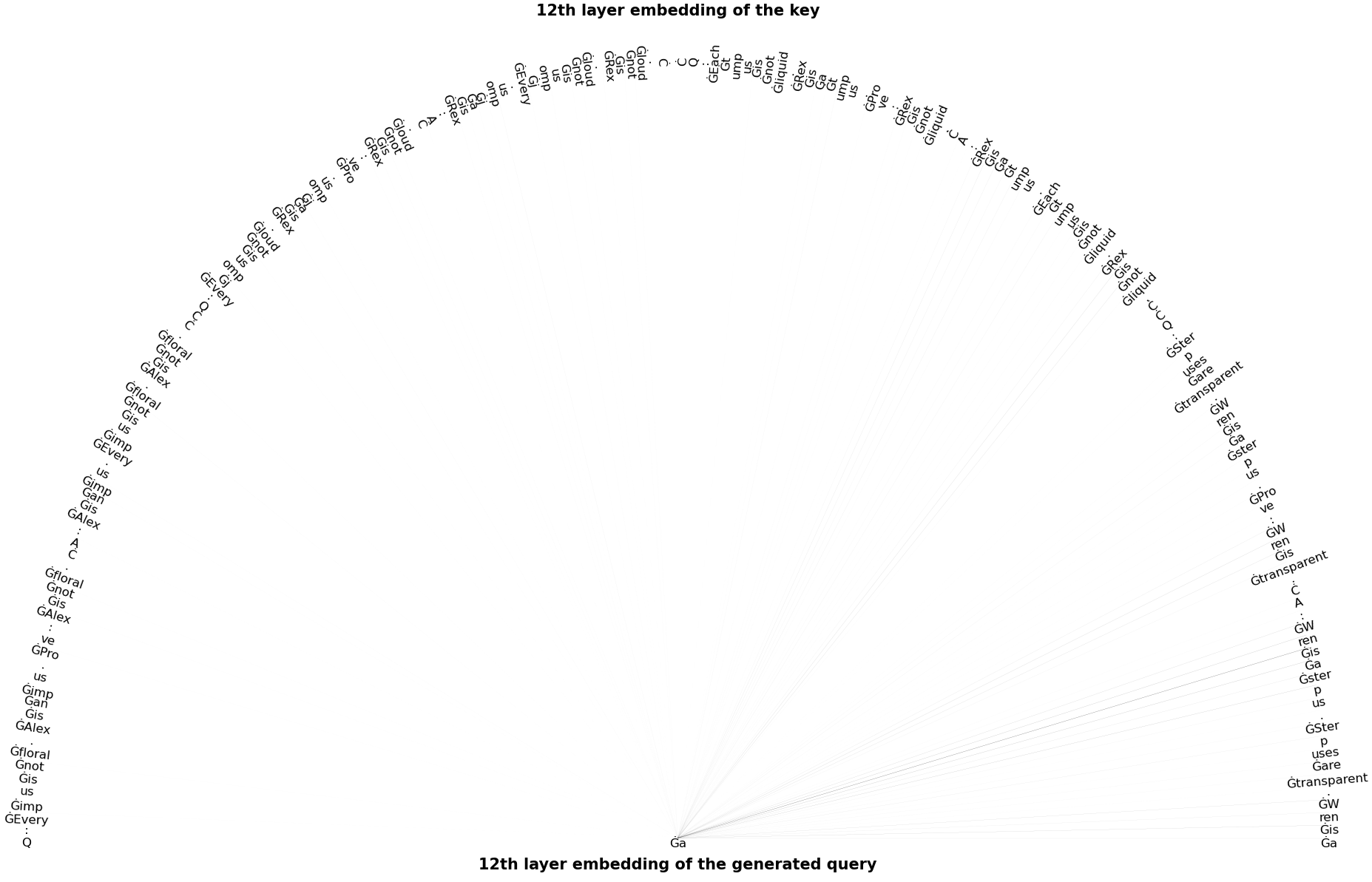}\label{fig:gpt2_H9}}
    \subfigure[The $\text{10}^\text{th}$ head]{\includegraphics[width=0.83\columnwidth]{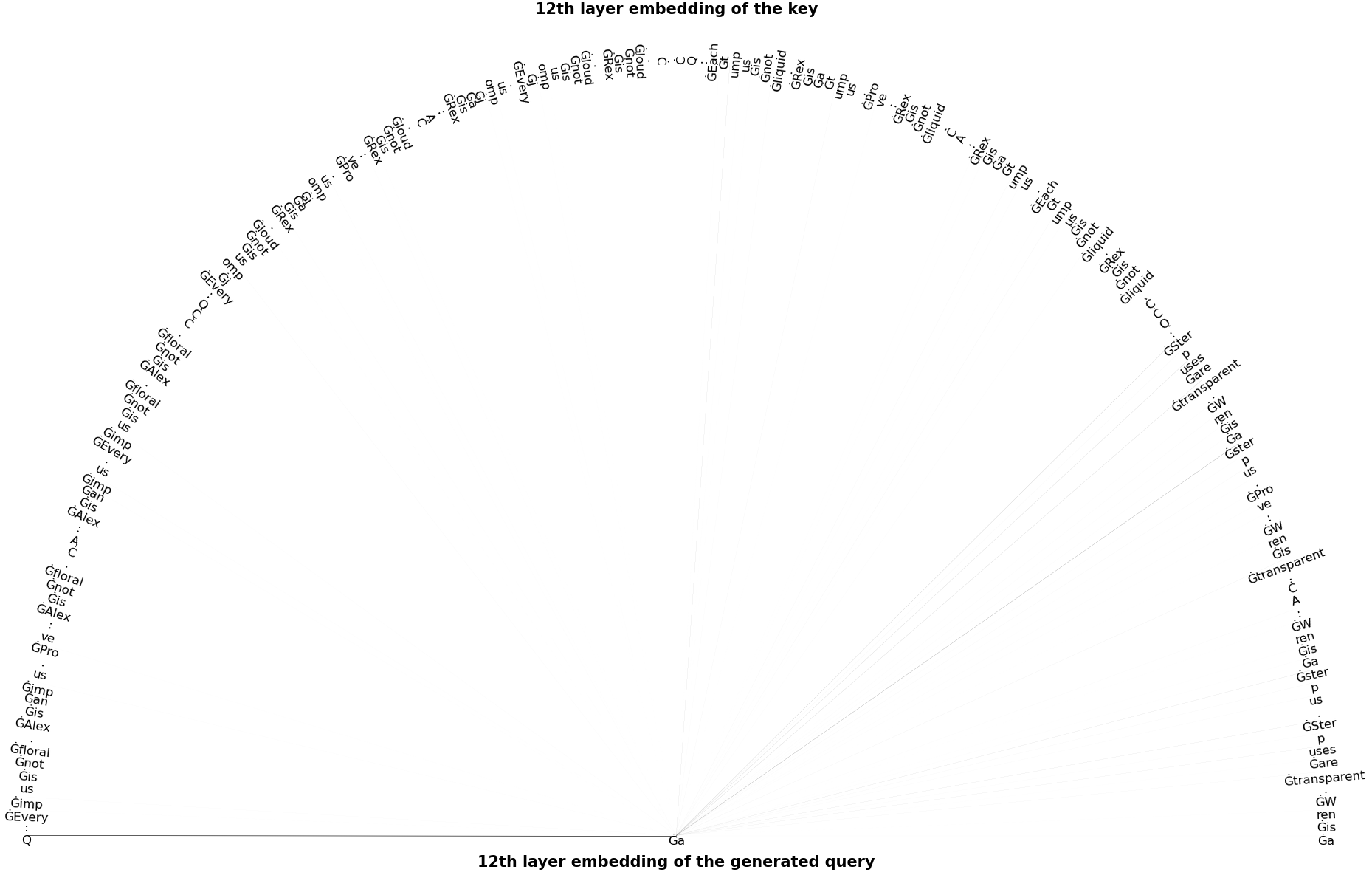}\label{fig:gpt2_H10}}
    \caption{Visualization of attention maps for each head in the final attention layer of the gemma2-9b-it model loaded with float16 precision. The model is prompted using the CoT prompt for the first proof of the implication elimination task from the PrOntoQA-OOD dataset \citep{saparov2023testing}, appended with the text: "Wren is a sterpus. Sterpuses are transparent. Wren is a". The thickness of each line connecting token embeddings is proportional to the corresponding attention score. The arc-shaped text represents the key embeddings, progressing from the bottom-left to bottom-right. The text at the bottom represents the query embedding of the last token.}
\label{Figure_ProntoQA_MP_attention_gpt2}
\end{center}
\end{figure}

\begin{figure}[!ht]
\begin{center}
\ContinuedFloat
\captionsetup{list=off,format=cont}
\centering
    \setcounter{subfigure}{10}
    \subfigure[The $\text{11}^\text{th}$ head]{\includegraphics[width=0.83\columnwidth]{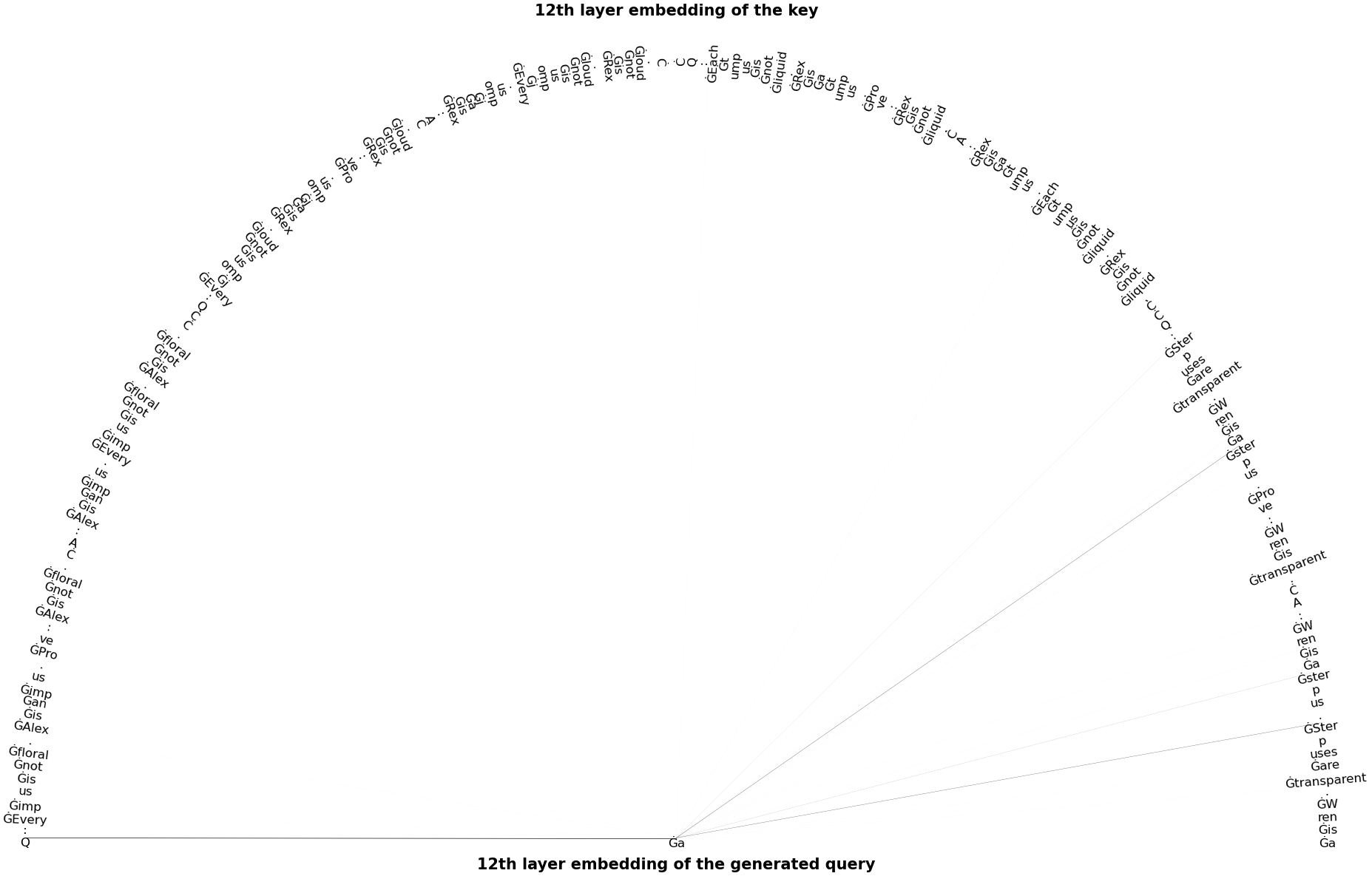}\label{fig:gpt2_H11}}
    \subfigure[The $\text{12}^\text{th}$ head]{\includegraphics[width=0.83\columnwidth]{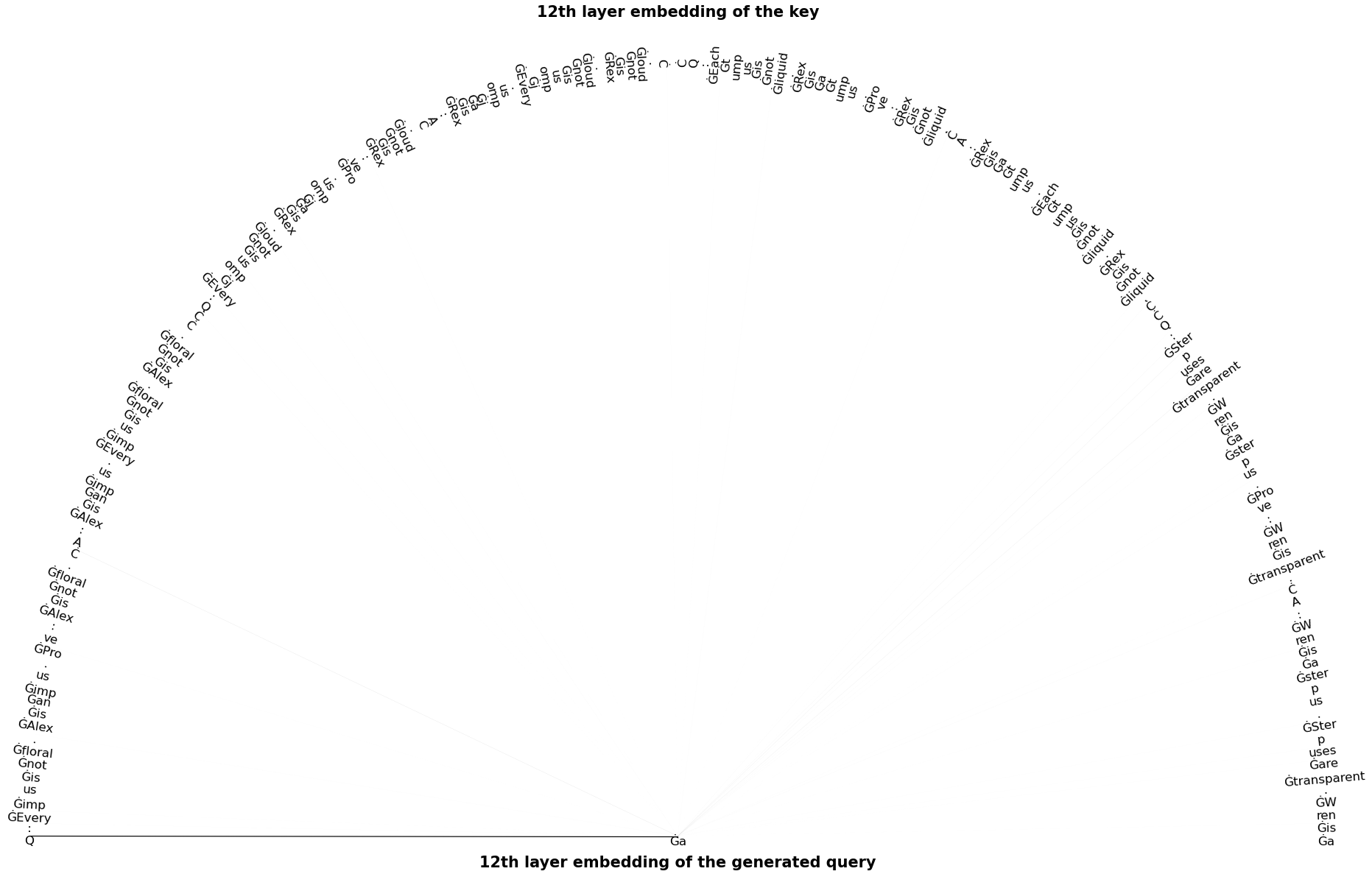}\label{fig:gpt2_H12}}
    \caption{Visualization of attention maps for each head in the final attention layer of the gemma2-9b-it model loaded with float16 precision. The model is prompted using the CoT prompt for the first proof of the implication elimination task from the PrOntoQA-OOD dataset \citep{saparov2023testing}, appended with the text: "Wren is a sterpus. Sterpuses are transparent. Wren is a". The thickness of each line connecting token embeddings is proportional to the corresponding attention score. The arc-shaped text represents the key embeddings, progressing from the bottom-left to bottom-right. The text at the bottom represents the query embedding of the last token.}
\label{Figure_ProntoQA_MP_attention_gpt2}
\end{center}
\end{figure}


\end{document}